\documentclass[]{report}
\usepackage[utf8]{inputenc}
\usepackage[T1]{fontenc}
\usepackage{minitoc}
\usepackage{tocbibind}
\usepackage{graphicx}
\usepackage{wrapfig}
\usepackage{array,multirow}
\usepackage{colortbl}
\usepackage{xcolor}
\usepackage{tabularx}
\usepackage{setspace}
\usepackage{tablefootnote}

\usepackage{fancyhdr}
\usepackage{hyperref}

\usepackage{microtype}

\dominitoc

\begin{document}

\begin{titlepage}
\center\footnotesize TUNISIAN REPUBLIC\\
\vspace{0.1em}\footnotesize MINISTRY OF HIGHER EDUCATION AND SCIENTIFIC RESEARCH\\
\vspace{0.1em}\footnotesize TUNIS EL MANAR UNIVERSITY\\
\vspace{0.1em}\footnotesize FACULTY OF SCIENCES OF TUNIS\vspace{0.1em}

\begin{figure}[!ht]
\centering
\includegraphics[width=0.26\textwidth]{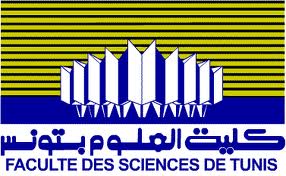}
\end{figure}

\vspace{2.5em}
{\bfseries\huge Master Thesis} \\
\vspace{0.25em}
{\slshape\normalsize Submitted for a} \\
\vspace{0.25em}
{\textrm{\normalsize\textbf{Research Master's Degree in Computer Science}}} \\ 
\vspace{1.9em}
{\slshape\normalsize By } \\
\vspace{1.9em}
{\bfseries\large Khouloud \textsc{Hwerbi}} \\
\vspace{3em}\hrule\vspace{0.5em}

{\textrm{\Huge\textbf{Ontology-Based Chatbot for Disaster Management: Use~Case CoronaVirus}}}
\vspace{0.5em}
\hrule

\vspace{4em}
\normalsize{\textbf{Defended on October 9, 2020 in front of the following jury:}} \\ \vspace{1em}

\begin{table}[!ht]
\centering
\resizebox{\linewidth}{!}{
\begin{tabular}{lll}
\textbf{President}&Hella \textsc{Kaffel Ben Ayed}&Associate Professor at FST\vspace{2px}\\
\textbf{Reporter}&Narjes \textsc{Doggaz}&Assistant Professor at FST\vspace{2px}\\
\textbf{Co-Supervisor}&Salvatore Flavio \textsc{Pileggi}&Lecturer at UTS\vspace{2px}\\
\textbf{Supervisor}&Sadok \textsc{Ben Yahia}&Full Professor at FST\vspace{2px}
\end{tabular}
}
\end{table}

\vspace{1mm}
\normalsize{\textbf{In the Laboratory: LIPAH}}

\vspace{4em}

\center\large{2019--2020}
\end{titlepage}

\chapter*{Abstract}

Today is the era of intelligence in machines. With the advances in Artificial Intelligence, machines have started to impersonate different human traits, a chatbot is the next big thing in the domain of conversational services. A chatbot is a virtual person who is capable to carry out a natural conversation with people. They can include skills that enable them to converse with the humans in audio, visual, or textual formats. Artificial intelligence conversational entities, also called chatbots, conversational agents, or dialogue system, are an excellent example of such machines.

Obtaining the right information at the right time and place is the key to effective disaster management. The term "disaster management" encompasses both natural and human-caused disasters. To assist citizens, our project is to create a COVID Assistant to provide the need of up to date information to be available 24 hours. With the growth in the World Wide Web, it is quite intelligible that users are interested in the swift and relatedly correct information for their hunt. A chatbot can be seen as a question-and-answer system in which experts provide knowledge to solicit users.

This master thesis is dedicated to discuss COVID Assistant chatbot and explain each component in detail. The design of the proposed chatbot is introduced by its seven components: Ontology, Web Scraping module, DB, State Machine, keyword Extractor, Trained chatbot, and User Interface.

\textbf{Keywords: } Chatbot, Conversational Agent, Ontology, Finite State Machine, Disaster Management, Covid-19

\thispagestyle{empty}
\setcounter{page}{0}
\chapter*{Dedication}
\vspace{7cm}

	\begin{flushright}
    \textit{To my parents} \\
	\textit{To my sister} \\
	\textit{To my friends} \\
	\textit{To all my family} \\	
	\end{flushright}

\thispagestyle{empty}
\setcounter{page}{0}
\chapter*{Acknowledgement}
\paragraph{I would like to thank Mr. Sadok \textsc{Ben Yahia}, Professor at the Faculty of Sciences of Tunis and director of the Laboratory of Computer Science in Programming, Algorithms, and Heuristics (LIPAH), for the trust he has accorded to me by agreeing to direct my master’s work. I thank him for his continuous availability and I would like to express my great and sincere gratitude to him;}

\paragraph{I would like to thank my co-director, Mr. Salvatore Flavio \textsc{Pileggi}, Lecturer at University of Technology Sydney for his support, for the ideas he’s given to me throughout this master’s thesis and sound and precious advice;}

\paragraph{I would like to express my gratitude to Mrs. Hella \textsc{Kaffal}, Professor at  the Faculty of Sciences of Tunis, for the honor she has done me by accepting to preside over the jury;}

\paragraph{I would also like to thank Mrs. Narjes \textsc{Doggaz}, Assistant Professor at the Faculty of Sciences of Tunis, for agreeing to join this jury as rapporteur;}

\paragraph{I would like to express my gratitude to Mr. Mohamed Taha \textsc{Bennani}, Assistant Professor at the Faculty of  Sciences of Tunis, for the help he gave me;}

\paragraph{I would also thank Ines \textsc{Osman}, a Ph.D. student at the Faculty of Sciences of Tunis, for the help she gave me throughout this master thesis.}

\thispagestyle{empty}
\setcounter{page}{0}

\newpage
\adjustmtc[3]
\tableofcontents
\listoffigures
\listoftables

\chapter{Introduction}
\minitoc
\newpage
The idea of building an artificial intelligent program dates back at least as far as Turing’s paper on “Computing Machinery and Intelligence” (Turing, 1950) was published, in which he asked the famous question “Can a machine think?”, and in the ongoing of the answer to this question, it can be said that the entire field of Artificial Intelligence (AI) has been enhanced.

Artificial intelligence (AI) is being utilized to improve technology development and applications due to its unbelievable capability of dealing with big data, complexity, high accuracy, and speedy processing. Marvin Lee Minsky, an American mathematician and computer scientist, one of the founding fathers of the science of artificial intelligence (AI), defined AI as “the science of making machines do things that would require intelligence if done by men.”

The AI assistants/chatbots have revolutionized by understanding human queries in different languages and appropriately extraction the meaningful information. The main aim of these chatbots is to supply informative, immediate, meaningful, and context-oriented responses to assist their users in the asked questions.

Chatbot is a computer program that interacts with users using natural Languages. Chatbot systems permit to realize simply a dialogue system based on natural language. chatbot can be used in multiple cases entertainment chatbots (Zeka \cite {spahic2019zeka}, museumBot \cite {varitimiadis2020towards}, etc.), educational chatbots (TutorBot \cite {lu2006using}, FreudBot~\cite {heller2005freudbot}, etc.), health care chatbots (SPeCECA \cite  {ouerhani2019spececa}, MedBot \cite  {modak2020voice}, Mandy \cite {ni2017mandy}, etc.), and banking service chatbots (\cite {dole2015intelligent}, \cite {altinok2018ontology}) and many more other domains.

The influence and the spreading of chatbots in society grow constantly, so we can take advantage of this aspect in an emergency situation like, the one which the whole world suffers these days, Covid-19 infection. The usage of chatbots could reduce the insufficient of real, correct, and availability of information provided to citizens. As machines do not need a break like human beings, they could ensure a 24-hour service for anyone who has access to a computer or to a smartphone. Citizens, therefore, does not need to conform to the opening hours of medical call centers.

Furthermore, chatbots are able to reply instantly and increase the accessibility of psychological support in such an emergency situation due to that Coronavirus is a new disease. Another benefit of using chatbots in emergency situations is the scalability of chatbots as they can respond to multiple users at the same time and can consequently reduce personnel costs.

The AI Chatbots retrieve information through different approaches. In modern practice, these approaches use a variety of repository structures such as conventional (Relational) and modern (NoSQL) database systems, ontologies, AIML, JSON files, etc.

In recent years, the development of ontologies has moved from the realm of artificial intelligence laboratories to the offices of experts in the field. Ontologies have become commonplace on the World Wide Web and a modeling trend in the development of information systems where we can take advantage of the great benefits they provide. Ontologies are comprehensive, human-readable, sharable, and formal it means that they are expressed in a language that has a well-defined semantics. Ontologies are important for application integration solutions because they enable a common and shared understanding of the data that exists within an application. Ontologies also make the communication between people and information systems easier.
 
In \cite {al2011ontbot}, \cite {nazir2019novel} and \cite {dobrila2009semantic} Knowledge of the Semantic Web was used to enhance the capabilities of a language-independent conversational agent-oriented towards solving question-answering tasks. Currently, developers who wish to use ontology repositories needs should learn about the contents of the ontology, which means understanding OWL or RDF, and to query these ontologies we need an ontology query language like SPARQL. Such requirements are some of the major causes, the Semantic Web has not become the main trend as fewer than expected users are utilizing such knowledge.

\section{Purpose}
In this master thesis, an ontology-based chatbot “COVID Assistant” is proposed to provide an easy to use, domain-independent, scalable, dynamic, and smart conversational agent.

The proposed system provides a text-to-text conversational agent to simulate a human conversation; the chatbot architecture integrates an NLP tool kit to extract and understand the user input, an ontology as a Knowledge-base, a finite state machine as the engine of the chatbot.

COVID Assistant Chatbot, developed in Python, covers all necessary and general information relevant to the Coronavirus disease, current status of contamination, trends, etc. For example, a chatbot user can ask: “what is the current status of Tunisia?” or “Tell me about the contamination status of the Corona Virus in Tunisia”, etc. To do so, we populated the ontology using web scraping technique on Covid-19 news from google web site to ensure that information provided to the user is always up to date, other information also are extracted from trusted sources.

\section{Structure}
This dissertation is organized as follows: 
\begin{enumerate}
    \item The  \textbf{next chapter} introduces the essential concepts for defining the field of study. indeed, a bibliographical study on the famous chatbots of the literature as well as a discussion is carried out on the different chatbots. Then, the foundation of the Semantic Web is studied. In the latter, we discussed the appearance of the semantic web, the ontological components and we cited the main languages of representation of ontologies. The last section is dedicated to the study of finite state machines.

 \item The \textbf{third chapter} presents the architecture of COVID Assistant and explains the work of each component separately.

 \item The \textbf{ fourth chapter} is devoted to the presentation of the conceptual study in which we studied the static and the dynamic view by defining a general use case diagram and a sequence diagram for each use case and an implementation section which describes technical details.

 \item The \textbf{ fifth chapter} is composed of two sections, the first one presents some screenshots of the execution of the chatbot and the second section contains a description of the validation step.
\end{enumerate}

\chapter{Background}
\minitoc
\newpage
\section{Introduction}
In this chapter, we start by talking about artificial intelligence, the field that includes chatbots, then we recall the appearance of chatbots, citing the most important ones in the literature up to today’s famous ones, then we start the foundation of the semantic web and the state machine. 
\section{Artifial Intelligence}
In computer science, artificial intelligence (AI), also called machine intelligence, is the intelligence demonstrated by machines, as opposed to the natural intelligence displayed by humans. According to AI textbooks, Artificial Intelligence is based on how a machine can perceive its environment and take actions that maximize its chance of successfully achieving its goals. In everyday language, the term "artificial intelligence" is often used to describe machines (or computers) that mimic the "cognitive" functions that humans associate with the human mind, such as "learning" and "problem-solving". AI gives the supreme power to simulate human behaving to a computer.

Chatbots are a real example of artificial intelligence, a chatbot is a conversational software system that is designed to emulate the communication capabilities of a human being that interacts automatically with a user. It employs Natural Language Processing and Pattern Recognition techniques to identify the meaning and provide meaningful responses to questions posed by humans.

It represents a modern form of customer assistance powered by artificial intelligence via a chat interface. For example, it makes it easy for customers to get responses to their queries in a convenient way without spending their time waiting in phone queues or send repeated emails.

\section{Chatbot history}
\subsection{Turing Test}
In 1950, a British mathematician, Alan Turing published a paper entitled “Computing Machinery and Intelligence” \cite {turing2009computing}  where he asked the famous question ”Can a machine think?” and he proposed the Turing test. It’s a test that somehow answers his question.
The fame of this question motivated the development of the first chatbot in 1966 ELIZA. And to answer the latter, in this paper Turing suggested ”The Imitating Game”. The Imitation Game can be played between three people: (A) who is a man, (B) who is a woman, and (C) who is the interrogator and who can be a man or a woman. The purpose of the interrogator is to determine who is the woman and who is the man.

Turing then wonders what will happen if A is replaced by a machine; can the interrogator tell the difference between the two? According to him, if the interrogator who converses with an unknown entity (via the keyboard, for example) thinks that this entity is human when the entity is in fact a computer, then this computer has managed to pass the Turing test.

ELIZA and PARRY are the first two programs that have passed the Turing Test. This article is widely regarded as one of the fundamental foundations of artificial intelligence and the Turing test which has been proposed in this article can be considered as a reference for assessing the intelligence of a computer system  \cite  {deshpande2017survey} . In 1990, an agreement was reached between Hugh Loebner and the Cambridge Center for Behavioral Studies to establish a competition based on the application of the Turing test.
\subsection{ELIZA}
\begin{wrapfigure}[]{r}{6cm}
\includegraphics[width=0.5\textwidth]{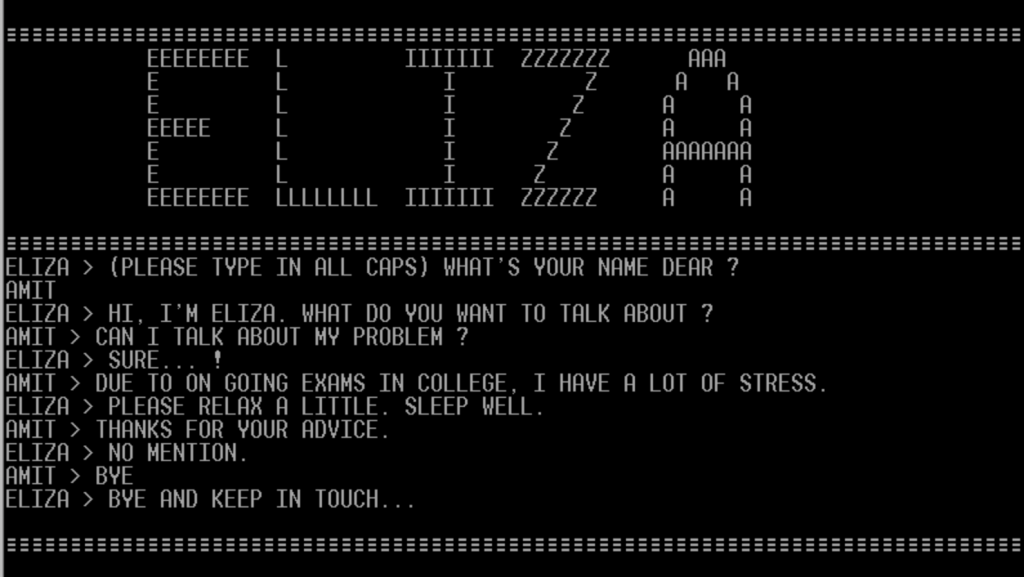}
\caption{ELIZA interface}
\end{wrapfigure}
ELIZA is the first conversational agent, developed between 1964 and 1966, by Joseph Weizenbaum. It is a system allowing human-machine interaction via natural language, it was designed to imitate a psychotherapist  \cite {weizenbaum1966eliza} . \newline
The program procedure is quite simple: ELIZA searches for text entered by the user and detects the presence of keywords. When such a word is found, it then applies pre-programmed rules to generate appropriate responses based on these keywords, otherwise, if a user does not enter a keyword, ELIZA responds with a default phrase, such as "Very interesting", "Please continue." or "Can you think of a specific example?" \newline
The mechanisms behind the transformation rule are a number of functions that serve to:

\begin{itemize}
\item Break down a character string according to certain criteria.
\item Re-assemble a decomposed chain according to certain specifications.
\end{itemize}

Considering the sentence "I am very unhappy these days" ELIZA's answer will be "How long have you been unhappy these days" applying the mechanism of transformation rules the chain is broken down into two parts, the first "I am" is replaced by "How long have you been" and the second part has remained stable. This transformation is in the form "I am BLAH" is transformed into "How long have you been BLAH", regardless of the meaning of BLAH. \newline
A slightly more complicated example is given by the sentence "It seems that you hate me" the answer will be "what makes you think I hate you" the program applied a decomposition into four parts (It seems that you, hate, me ). The transformation will be to throw out the first part, translate the two known words ("you" into "me" and "me" into "you"), add a common phrase "what makes you think" and keep the third part. \newline Although using a simple technique, ELIZA was able to deceive a large number of people who believed that the program really listened to their problems. Here is an example of a conversation with ELIZA:
\begin{center}
\begin{table}[!ht]
\centering
\begin{tabular}{|c|}
\hline
\begin{minipage}{11cm}
\medskip \par
\begin{itemize}
 \item User: Men are all alike.
	\item ELIZA: IN WHAT WAY?
	\item User: They're always bugging us about something or other.
	\item ELIZA: CAN YOU THINK OF A SPECIFIC EXAMPLE?
	\item User: Well, my boyfriend made me come here.
	\item ELIZA: YOUR BOYFRIEND MADE YOU COME HERE
	\item User: He says I'm depressed much of the time.
	\item ELIZA: I AM SORRY TO HEAR YOU ARE DEPRESSED.
	\item User: It's true. I am unhappy
	\item ELIZA: DO YOU THINK COMING HERE WILL HELP YOU NOT TO BE UNHAPPY
	\item User: I need some help, that much seems certain.
	\item ELIZA: WHAT WOULD IT MEAN TO YOU IF YOU GOT SOME HELP
	\item User: Perhaps I could learn to get along with my mother.
	\item ELIZA: TELL ME MORE ABOUT YOUR FAMILY
 \item \ldots
\end{itemize}
\medskip
\end{minipage} \\
\hline
\end{tabular}
\caption{Typical conversation with ELIZA \cite {weizenbaum1966eliza}}
\end{table}
\end{center}
ELIZA generally asked users questions to raise them and minimize the risk of error. This type of approach is frequently used in chatbots to make them look more human. But, being a rules-based system, ELIZA could not have meaningful conversations with humans, but rather it could deceive a human into giving the illusion that it is not a computer program.
\newpage
\subsection{PARRY}
After ELIZA, Kenneth Colby, a psychiatrist, developed PARRY at Stanford University. PARRY was written to imitate a person with paranoid schizophrenia. PARRY was tested in the early 1970s using a variant of the Turing test. A group of experienced psychiatrists analyzed a combination of real patients and computers running PARRY via teletypewriter. Psychiatrists were only able to make the correct identification 48\% of the time 
\cite  {khante2019survey}. \newline Despite everything, Parry is still rule-based and has a response model structure similar to ELIZA, but it has an advanced control structure, language comprehension capabilities and above all a mental model that can simulate emotions of the bot like anger, fear\ldots For example, Parry will respond with hostility if the level of anger is high \cite {shum2018eliza}.
PARRY’s goal was not to help mental health therapy; but rather to show how technology could be used to imitate a person with mental health problems \cite {stiefel2018chatbot}
\subsection{A.L.I.C.E}
\begin{figure}[!ht] 
\begin{center}
\includegraphics[width=11cm]{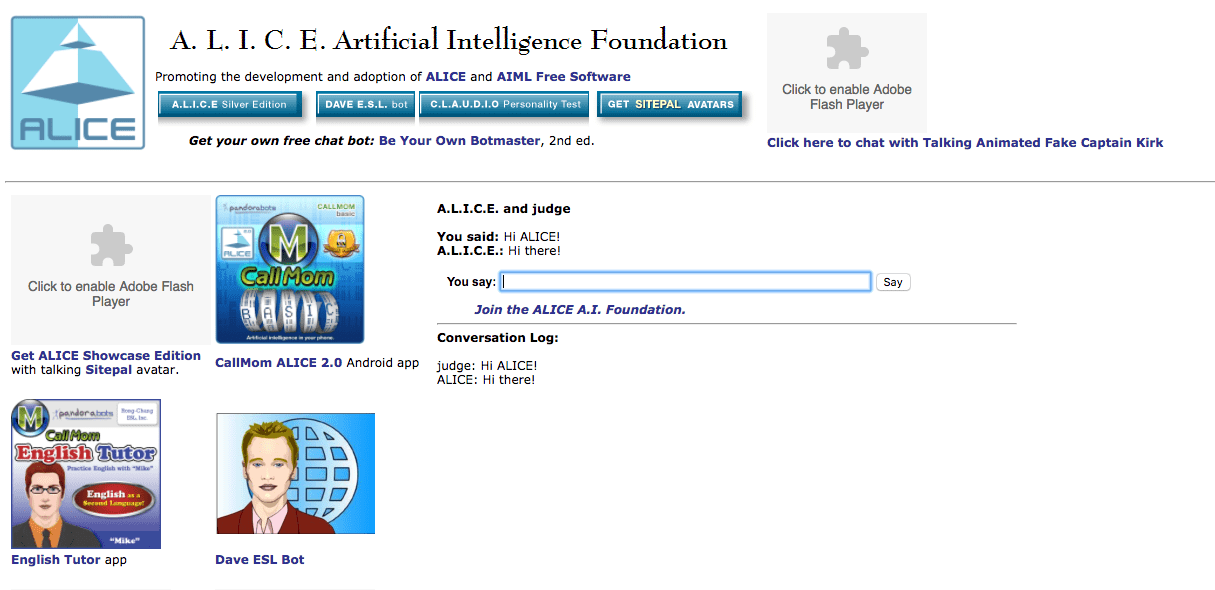}
\caption{A.L.I.C.E interface}
\end{center}
\end{figure} 
Artificial Linguistic Internet Chat Entity or ALICE, developed in 1995 by Richard Wallace, is an open source natural language processing chatbot, inspired by ELIZA. Richard Wallace, has hand-written a database of thousands of possible conversational gambits.

ALICE uses AIML (Artificial Intelligence Mark-up Language), an XML-like language designed to create stimulus-response chat robots. The ALICE knowledge base is made up of question and answer modules, called categories, and structured with AIML. The learning model in ALICE is called supervised learning because a person, called a botmaster, plays a crucial role. In fact, the botmaster monitors robots’ conversations and creates new AIML content to make responses more appropriate, precise or believable \cite {al2011ontbot}.\newline
Each category includes a response model and a set of conditions that give meaning to the model known as context. Then the model prepares it and compares it to the nodes of the decision tree. When user data is associated, the chatbot responds or performs an action. AIML models repeat user input using recursive techniques and these are not always meaningful responses. Therefore, rules based on character strings are necessary to determine whether the response creates a correct or meaningful response \cite  {nuruzzaman2018survey}. The authors of \cite  {shah2006alice} consider ALICE to be the head and the shoulders above the other conversational agents. ALICE was the recipient of Loebner’s annual instantiation of the Turing test for artificial intelligence three times in 2000, 2001, and 2004.
\newline As this is a predefined set of questions and answer questions, it cannot satisfactorily answer each question \cite  {thorat2020review}.
There are several notable research studies where ALICE has been used. The first was an English and German conversational partner for Chinese students [21]. However, a large proportion of students did not like the chatterbot responses and made poor comments on the system \cite  {schumaker2009interaction}. \newline 
Despite all of its drawbacks, none of today's chatbots would have been possible without the revolutionary work of Dr. Wallace. Additionally, Wallace's bot served as inspiration for the associated operating system in Spike Jonze's 2013 sci-fi romance film Her \cite {shaikh2019survey}.
\begin{center}
\begin{table}[!ht]
 \centering
\begin{tabular}{|c|}
\hline
\begin{minipage}{11cm}
\medskip \par
$<$aiml version=1.0.1 encoding=UTF-8$ >$ \\
$<category>$ \\
$<pattern>$ HELLO BOT $</pattern> $\\
 $<template>$ \\
$<random>$ \\
 $<li>$ Hi! Nice to meet you$ </li> $\\
 $<li>$ Hello, How are you?$ </li>$ \\
 $<li>$ Hello! $</li>$ \\
$</random>$ \\
$</template> $\\
$</category> $\\
$</aiml> $
\medskip
\end{minipage} \\
\hline
\end{tabular}
\caption{Example of ALICE's AIML structure \cite {trivedi2019chatbot}}
\end{table}
\end{center}
Richard Wallace created Alice, and it was launched in November 1995, it was rewritten in 1998 in Java language.
\newpage
\subsection{Mitsuku}
\begin{wrapfigure}[]{l}{6cm}
\includegraphics[width=0.5\textwidth]{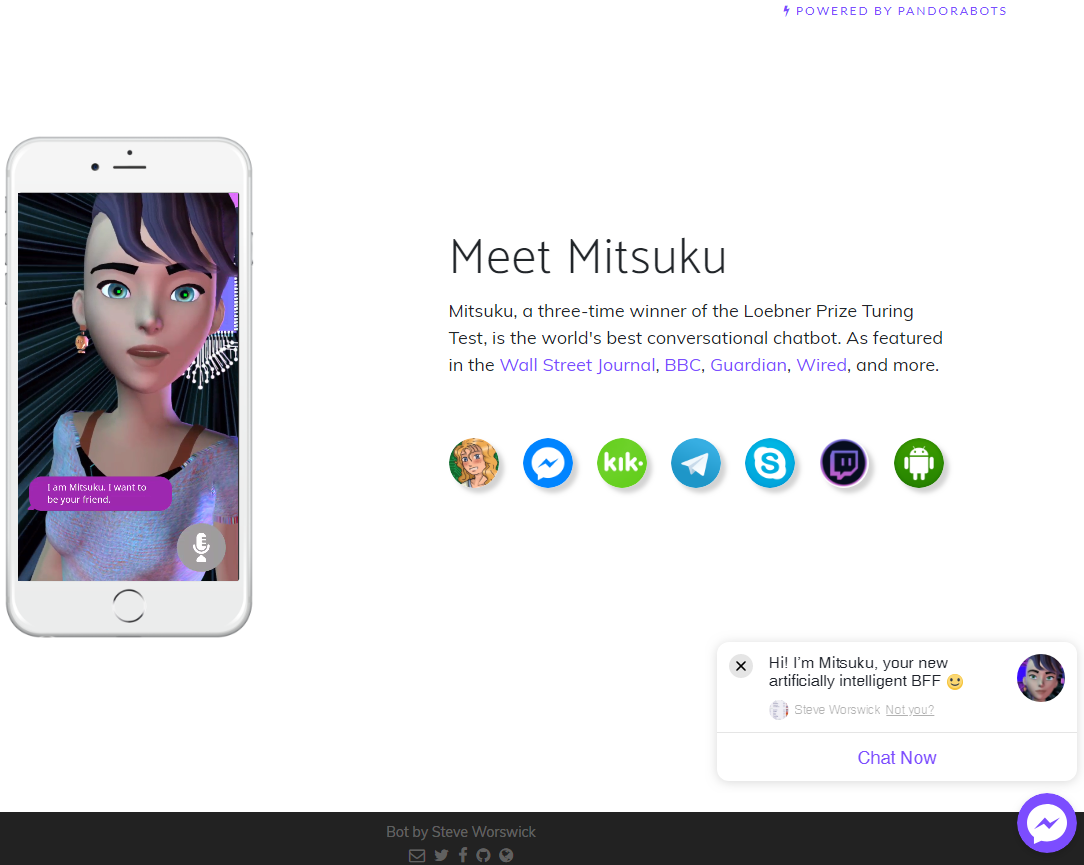}
\caption{Mitsuku}
\end{wrapfigure}
Mitsuku is a chatbot created by Steve Worswick. He has won the Loebner Prize five times (in 2013, 2016, 2017, 2018 and 2019). It was designed for general typed conversation based on rules written in AIML.

Mitsuku can hold a long conversation, learn from the conversation, remember personal details about the user (age, location, gender, etc.). Its functionality includes the possibility of reasoning with specific objects. For instance, if someone asks, "Can you eat a car? Mitsuku searches for the properties of "because" finds that the value of "made from" is set to "steal" and responds to "no" because a car is not edible \cite {sharma2020comparative}.
Mitsuku is a multilingual bot and uses supervised machine learning. As he learns something new, the data is sent to the human manager for verification. Only verified data can be integrated and used by the application~\cite {nuruzzaman2018survey}. Mitsuku is available as a flash game on Mousebreaker Games and Facebook Messenger, Twitch group chat, Telegram, and Kik Messenger under the username "Pandorabots".

\subsection{IBM's Watson}
\begin{wrapfigure}[]{l}{6cm}
\includegraphics[width=0.4\textwidth]{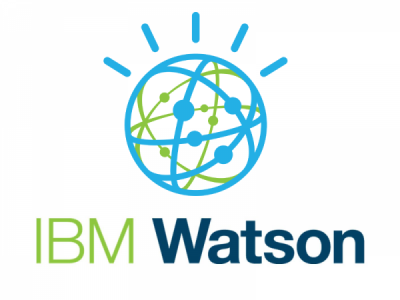}
\caption{IBM Watson}
\end{wrapfigure}
Watson\footnote{\url{https://www.ibm.com/watson }}, is a rules-based conversational agent, developed in 2006 by an IBM team as part of the DeepQA project. The Agent is a Question-Answer framework was designed specifically to win the American TV show Jeopardy. What he did in 2011.

Since then, IBM Watson has offered services to create chatbots for different domains that can process large amounts of data \cite {bhagwat2018deep}.
Watson was designed to apply advanced technologies for natural language processing, information retrieval, knowledge representation, automated reasoning, and machine learning. Watson integrates a variety of technologies, including Hadoop, the Apache Unstructured Information Management Architecture (UIMA) framework to examine the sentence structure and grammar of the question to better assess what is being asked \cite {nuruzzaman2018survey}.

\subsection{Apple's Siri}
\begin{wrapfigure}[]{l}{7cm} \includegraphics[width=0.5\textwidth]{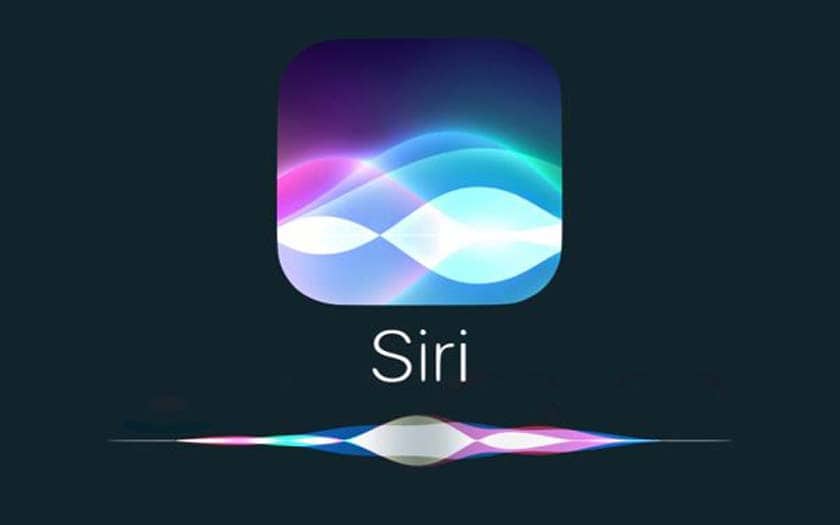}
\caption{Siri}
\end{wrapfigure}
Apple released Siri\footnote{\url{https://www.apple.com/siri/}} in 2011. Since then, several Intelligent Personal Assistants (IPAs) have been built and introduced to the market. Siri is a virtual assistant specifically available on Apple products and has access to Apple applications. \newline
Siri can read user emails, text contacts, change music, make phone app calls, find restaurants, find books, set alarms, and give directions. The user can dictate an SMS to send, a search to be done on the web with Safari or even he can chat with it. The genre, accent, and language of Siri are configurable and changeable. Siri uses ASR (Automatic speech recognition) to translate human speech into text. Using natural language processing (part of speech tagging, noun-phrase chunking, dependency, and constitute parsing), it translates the transcribed text into ``parsed-text". Using question and intent analysis, it analyzes the parsed text and detects user commands and actions.\newpage
\subsection{Amazon's Alexa}
\begin{wrapfigure}[]{l}{7cm}
    \includegraphics[width=0.4\textwidth]{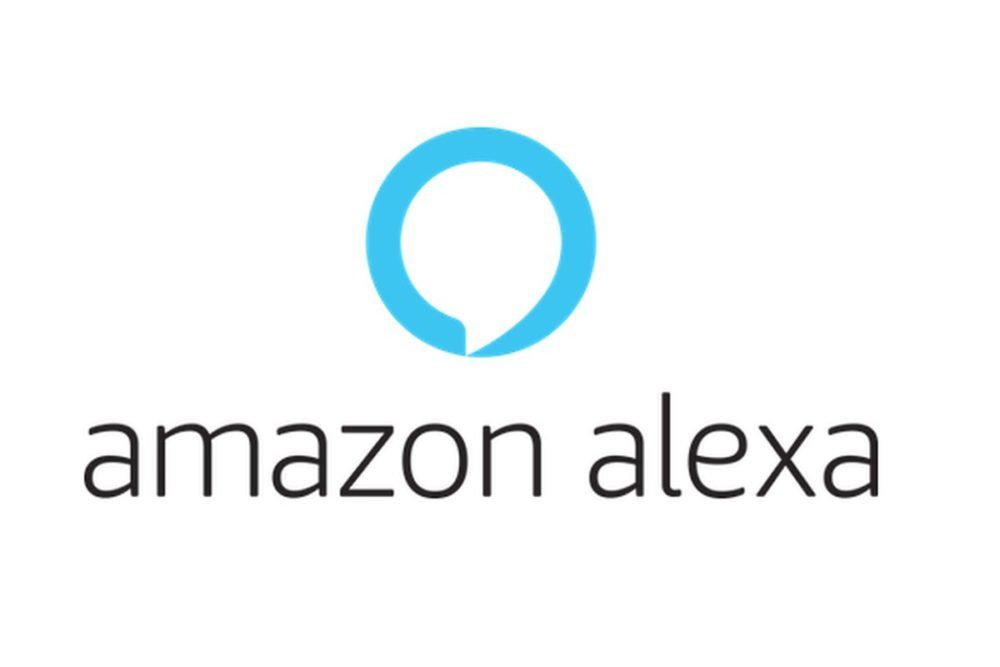}
  \caption{Amazon Alexa}
\end{wrapfigure}
Amazon Alexa\footnote{\url{https://fr.wikipedia.org/wiki/Amazon\_Alexa}}, also known as Alexa, is a virtual assistant AI technology developed by Amazon in November 2014, first used in Amazon Echo smart speakers and then Alexa became the voice service that powers many Amazon devices like Dot, Spot, Show, and Amazon Fire TV. It provides capabilities and skills that allow customers to interact with devices more intuitively using voice. 
Alexa uses natural language processing algorithms for voice interaction. It uses these algorithms to receive, recognize, and respond to voice commands
\cite {khante2019survey}. It is capable of voice interaction, reading music, making to-do lists, setting alarms, reading podcasts and audiobooks, and giving the weather, traffic, and other information in real-time.
Currently, interaction and communication with Alexa is available in English, French, Spanish, German, and Japanese.

\subsection{Microsoft's Cortana }
\begin{wrapfigure}[]{l}{6cm}
\includegraphics[width=0.5\textwidth]{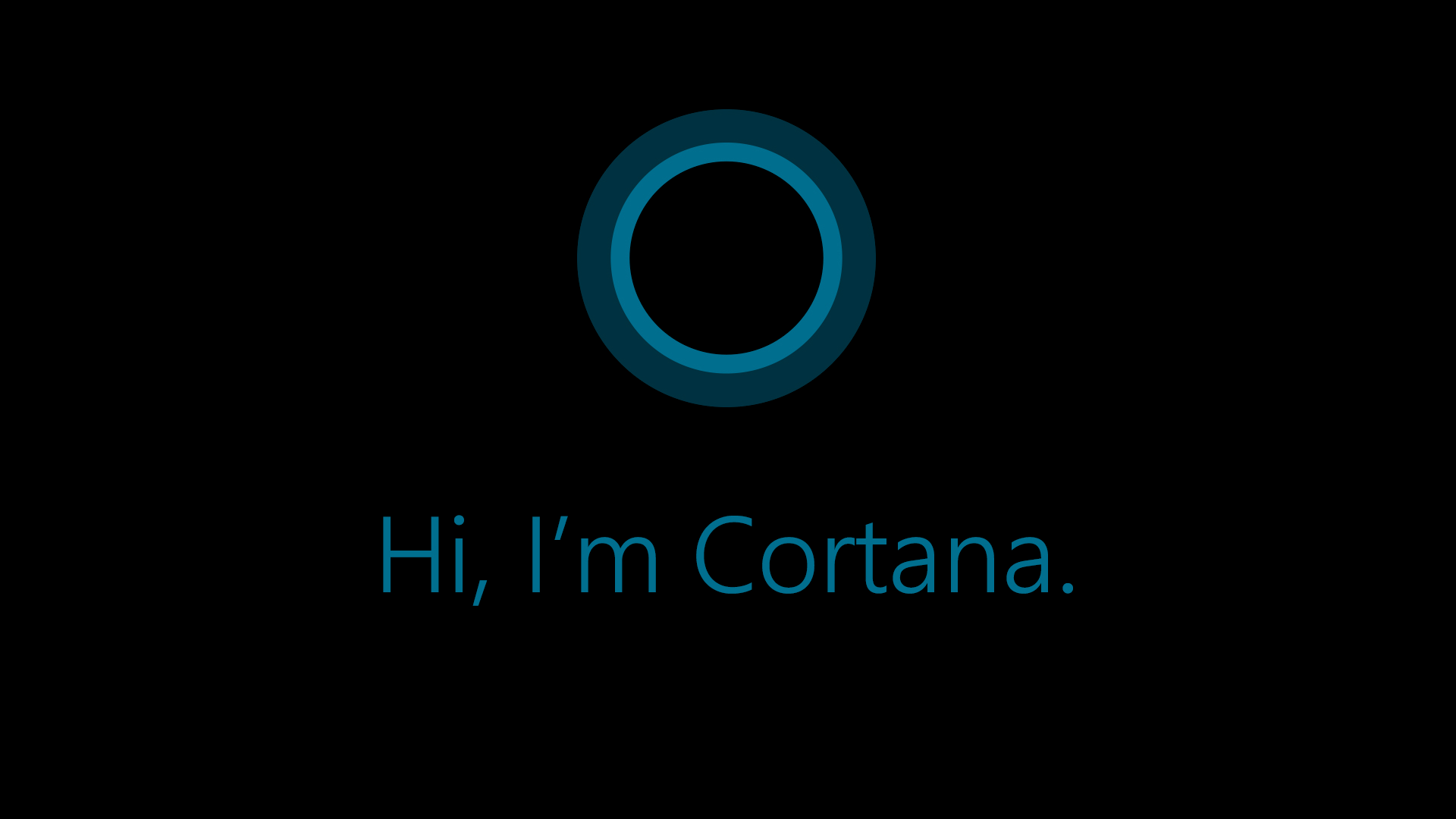}
\caption{Cortana }
\end{wrapfigure}
Cortana\footnote{\url{https://support.microsoft.com/en-us/help/17214/cortana-what-is}} is the name of the intelligent personal assistant and knowledge browser for Windows Mobile and Windows 10. Cortana was introduced for the first time at the Microsoft BUILD Developer Conference in April 2014 in San Francisco. He is named after Cortana, a synthetic intelligence character from Microsoft's Halo video game franchise, written in C\#. \newline
There are two ways to use Cortana: you can use voice commands or you can type your commands from the Start menu. Cortana can be utilized for basic search functions (e.g. weather), can access users' Google calendars, read Outlook emails, give travel time estimates, provide directions and integrate to OneNote to display user ratings. 

\subsection{Google assistant}
\begin{wrapfigure}[]{l}{5cm}
\includegraphics[width=0.4\textwidth]{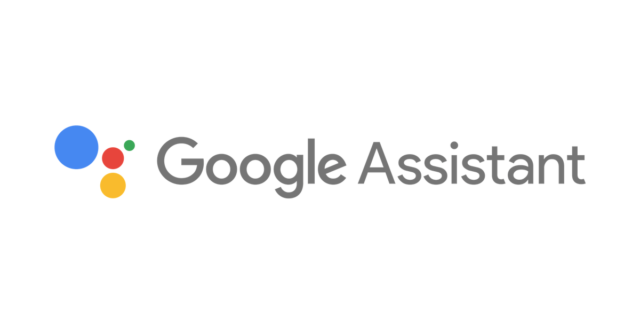}
\caption{Google Assistant}
\end{wrapfigure}
Google Assistant\footnote{\url{https://assistant.google.com/explore}} is Google's artificial intelligence announced at the Google I/O conference in May 2016. In line with Alexa, the assistant created by Amazon, or Apple's Siri, Google Assistant is not the first attempt Google on the subject. Its predecessor, Google Now, already made it possible to make requests via voice recognition on compatible devices.
Google Assistant, unveiled in 2016, introduces a conversational dimension, the assistant being able to maintain a dialogue with the user and understand natural language. It is able to recognize up to six different voices and associate them with their respective Google account. The Google Assistant is built into most Android 5.0 and above smartphones and can communicate with the user via voice or IM-like interaction. It is also found in the Google Home range, Google Allo, in some versions of Android Wear Since then, Google has extended the presence of Google Assistant to its system for connected watches, Wear OS by Google, and Android TV, its interface for connected TV. The assistant manages simple requests (weather, day schedule ...), can answer questions, tell stories, or give information on traffic or opening hours of shops or administrations.
\subsection{Discussion}
In this section, we provide a study of chatbots construction and enhancement methods. We first introduce three chatbot designs like in \cite {adiwardana2020towards} and \cite {peng2019survey}, which are: rule-based, retrieval-based and generation-based methods, followed by second way of decomposing that is proposed in \cite {schaub2019systemes} and \cite {almansor2019survey} they specify two categories of conversation agents, namely task-oriented and non-task-oriented. Finally, we present a comparative study conducted by authors in \cite {sharma2020comparative} in which they rank eight chatbot according to their performance in i) Assessment of Factual Questions, ii). Assessment of Conversational Attributes, and iii). Evaluation of Exceptional queries. 

\begin{figure}[!ht] 
\includegraphics[width=14cm]{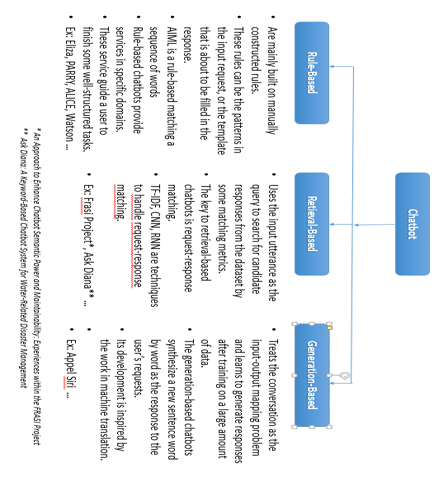}
\caption{Chatbot design}  
\label{design}
\end{figure}

The image \ref{design} explains the first proposed decomposition and gives some example for each design. 
\newpage

The second classification goal-oriented or non-goal-oriented is now presented. 

\begin{itemize}
\item A goal-oriented system is one that is designed to help users achieve their goals or complete a specific task in a specific domain such as making a booking, education, healthcare, shopping, or ordering food. They are focusing on developing natural language understanding (NLU) methods which parse the utterance from the user into predefined semantic.
\item A non-goal-oriented dialogue system provides users with the means to participate in different domains such as a game, chitchat, or entertainment, without providing the user the help to complete any task in a specific job. An example of this system is the chatbot which chats with the user in a similar way to a human and provides reasonable and relevant responses ELIZA, PARRY. 
\end{itemize}

Authors in \cite{sharma2020comparative} evaluated eight chatbots in the literature based on their performance in all the tests carried out, namely i) Assessment of Factual Questions, ii) Assessment of Conversational Attributes and iii) Assessment of Exceptional Queries. The results are as follows
\begin{center}
\begin{table}[!ht]
\centering
\begin{tabular}{|l|}
 \hline
Rank  \\
\hline
1. Mitsuku \\
\hline
2. Google Assistant  \\
\hline
 3. Siri    \\
\hline
 4. A.L.I.C.E.   \\
\hline
5. Rose   \\
\hline
6. Jabberwacky         \\
\hline
7. Machine Comprehension Chat Bot \\
\hline
8. Eliza    \\
\hline

 \end{tabular}
 \caption{Rank List \cite {sharma2020comparative} }
\end{table}
\end{center}
Mitsuku performed the best on all three parameters. It provided appropriate responses and scored better on all parameters than the other chatbots included in the study. Mitsuku could access factual information about the query and it was able to carry the conversation by starting new topics and handled insults appropriately. Google Assistant was reasonably good in all the assessments viz factual questions, conversational attributes, and exceptional queries. In factual questions, it was quite descriptive. In conversational attributes, it conversed well and was quite informative as it can surf the internet. And in exceptional queries, its performance as well and ranked just below Mitsuki.  Even though Siri fared well in all the three parameters, it still lagged behind Google Assistant and Mitsuku in the final averages. This underlines the fact that Siri is mainly designed to be a virtual assistant and not a fully functional chatbot, which explains its delay in the parameter "Ability of the chatbot to launch new topics". ALICE fares well in all the metrics but lacking in the ability to access chat history but like, Mitsuku, it performed the best in the Assessment of factual questions parameters and satisfactorily in the Assessment of Exceptional Queries parameter, as compared to other chatbots included in the study and below its rank.
Eliza, an early chatbot designed in 1966 at the MIT Artificial Intelligence Lab, imitates a Rogerian Psychotherapist. A final comparative analysis founds Eliza to be the weakest performing chatbot out of all the chatbots included in this study.

\section{Foundation of the Semantic Web}
\subsection{Concept of the semantic web}
The current Web is a set of documents dedicated to humans, stored and manipulated in a purely syntactic way. So, the current Web cannot be manipulated in an intelligent way by computer programs because they are human-oriented.

The growing interest in searching for information on the Web has given rise to the Semantic Web initiative. The Semantic Web, also known as the Intelligent Web, is a body of knowledge where all machines can semantically link data on the Web, thus understanding their meanings, and access them more intelligently, to improve the dialogue between applications and the interaction with the user by providing a better quality of search tasks. It can also be seen as a supplementary layer of knowledge above the current Web or an extension of the current Web.

To reach the goal of creating an intelligent web that will allow reporting on semantics through the handling of the meaning of information through ontologies... Ontologies propose semantic representations of knowledge that can be manipulated by machines.

The main objective of ontologies is to provide the Web with a semantic layer that ensures information retrieval at both the syntactic and semantic levels \cite {zghal2010contributions}.

\subsection{Ontology}
\subsubsection{Review of ontology definitions}
The word ontology is the union of two terms ”onto” and ”logy”. The first term ”onto” (or ”ontif”) means ”I am”. The second term ”logy” (or ”logos”) is a Greek suffix meaning ”science” or ”speech”. In the philosophical field, Aristotle defined ontology as the science of Being \cite {zghal2010contributions}. 

For the past decade, computer scientists have been using the term "Ontology",  The definition proposed by Gruber 1993 in \cite {gruber1993translation} "An ontology is an explicit specification of a conceptualization".

One of the first definitions was given by \cite{neches1991enabling} "An ontology defines the basic terms and relations comprising the vocabulary of a topic
area as well as the rules for combining terms
and relations to define extensions to the
vocabulary".

In \cite {studer1998knowledge} "An ontology is a formal, explicit specification of a shared
conceptualization" 
\begin{itemize}
 \item A `conceptualization' refers to an abstract model of some phenomenon in the world
by having identified the relevant concepts of that phenomenon
\item `Explicit' means that the type of
concepts used, and the constraints on their use are explicitly defined. 
\item `Formal' refers to the fact that the ontology should be
machine-readable, which excludes natural language.
\item `Shared' reflects the notion that an ontology
captures consensual knowledge, that is, it is not private to some individual, but accepted by a group. 
\end{itemize}

\subsubsection{Compostion} 
Despite the diversity of ontological models, they are based on the following main concepts
following \cite {barkat2017utilisation}:
\begin{itemize}
 \item \textbf{Concepts (classes or entities). } A concept can represent a material object, a notion or an idea.
 A concept can be divided into three parts: one or more terms, a notion, and a set of objects. The intension contains the semantics of the concept. It is expressed in terms of properties and attributes, rules, and so on.
and constraints. The set of objects is also called an extension of the concept \cite {zghal2010contributions}.
 \item \textbf{Instances (individuals or objects). }Constitute the extended definition/extension of concepts, they represent all the individuals in the ontology. Instances can be linked by conceptual relations of identity and difference.
 \item \textbf{Attributes (properties).  }Properties are used to define links for individuals in the domain of ontology. They describe and characterize instances of ontology concepts with characteristic values or associations with other concepts.
 annotation property, data property, object property.
\item \textbf{The relations. }These are the defined associations between concepts in the ontology, such as, for example, the "is-a" subsumption relationship that is used to organize concepts into a hierarchy.
\item \textbf{Axiomes: }The latter refer to statements and assertions accepted as true in the field of ontology. They are used to verify the consistency of the ontology and to infer new knowledge.
they are entity-related assertions. Instead of relying on entity labels and terms to convey semantics, the designer of ontologies must constrain the possible interpretation of entities through the judicious use of logical axioms to make their meanings much more precise.
\end{itemize}
\subsection{Representation languages}
In the literature, several languages have been used for the description of ontologies. These languages offer different levels of expressiveness \cite {zghal2010contributions}.
\begin{itemize}
\item \textbf{Extensible Markup Language\cite  {rose2010automatic} (XML). } is a simple text-based format for representing structured information: documents, data, configuration, books, transactions, invoices, and much more. It was derived from an older standard format called SGML (ISO 8879), in order to be more suitable for Web use.

It does not allow to impose semantic constraints on the meaning of the described documents. An XML schema is a description of the type of an XML document. The schema contains a set of rules. These rules are constraints on the structure and content of the XML document. An XML document must respect the XML schema dedicated to it in order to
to ensure the validity of the document according to its scheme.
The DTD (Document Type Definition) is used to define a grammar to check the conformity of the XML document.
The DTD and the XML schema are developed to express XML schemas.
 \item \textbf{Resource Description Framework (RDF) \cite {klyne2009resource}. } is an infrastructure that enables the encoding, exchange, and reuse of structured metadata. RDF additionally provides a means for publishing both human-readable and machine-readable vocabularies designed to encourage the reuse and extension of metadata semantics among disparate information communities. To identify resources, RDF uses URIs (Uniform Resource Identifiers). RDF declarations take the form of triplets : (\textit{i}) a subject designating the resource to be described, (\textit{ii}) a predicate representing an attribute or a property applied to the resource, and (\textit{iii}) an object representing the value of the property. RDF triplets can be represented graphically through a graph or described in XML. RDF has been extended by a set of manufacturers giving rise to the RDF-Schéma (or RDFS) model.
 \item \textbf{RDF scheme (RDFS) \cite {brickley1999resource}. } RDFS allows us to build concepts, possibly defined from other concepts, shared on the web. RDFS expressions are written in the form of RDF triplets. It provides mechanisms for the description of similar resources and the relationships between these resources. The descriptions of the RDF scheme RDFS vocabulary are written in RDF. The two RDF languages
and RDF schema are referenced by RDFS. The RDFS has mechanisms for the
description of the resources and their characteristics.
 \item \textbf{Darpa Modeling Language of Ontology
+ Ontology Inference Layer (DAML+OIL) \cite {connolly2001daml+}.} is a semantic markup language for Web resources. It builds on earlier W3C standards such as RDF and RDF Schema and extends these languages with richer modeling primitives. DAML+OIL provides modeling primitives commonly found in frame-based languages. DAML+OIL (March 2001) extends DAML+OIL (December 2000) with values from XML Schema datatypes. DAML+OIL was built from the original DAML ontology language DAML-ONT (October 2000) in an effort to combine many of the language components of OIL. The language has a clean and well-defined semantics
this is a fusion between two languages DAML and OIL.
The Darpa Modeling Language of Ontology aims to provide the foundation for the generation of the Semantic Web. The language first adopted RDF / RDFS as an ontology language to ensure semantic interoperability. As RDF(S) not sufficiently expressive, with respect to the requirements of the future Web, a new language was adopted named DAMLONT. The latter represents an extension of RDF(S) with the capabilities of a knowledge representation language. In the same period, a group of European researchers developed an ontology language called OIL. This language has a syntax based on RDF. It has been built in such a way that its semantics can be specified through a description. very expressive logic. 
\item \textbf{Ontology Web Language (OWL) \cite {smith2004owl}. } It enriches the
of the RDF schema by defining more vocabulary for the description of complex ontologies. The OWL language is based on formal semantics denied by a rigorous syntax.  The OWL language is a language for the representation of ontologies within the Semantic Web. It is an RDF-based language.  The OWL language is derived from the ontology language DAML+OIL. OWL allows, in addition to RDF(S) primitives, relations between classes (disjunction, intersection, union, etc.), equality, cardinality,  etc. The OWL language allows the definition of property types (object property, annotation property, data property), property characteristics, and enumerated classes. The OWL ontology language is split into three sub-languages with an ascending expressivity, namely: OWL-Lite, OWL-DL (Ontology Web Language-Description Logic) and OWL-Full.
\end{itemize}
\section{Finite State Machine}
Finite State Machine have been used in the construction of conversational agents workflow for example the work of \cite {yi2017chatbot}, \cite {abowd1995formal} and \cite {raux2009finite}.

A finite state machine contains a finite number of states and
produces outputs on state transitions after receiving inputs. Finite
state machines are widely used to model systems in diverse areas,
including sequential circuits, certain types of programs, and, more
recently, communication protocols. 

A finite state machine contains a finite number of states and produces outputs on state transitions after receiving inputs. Finite state machines are widely used to model systems in diverse areas, including sequential circuits, certain types of programs, and, more recently, communication protocols.
In the 1960s and early 1970s, several researchers proposed models to explain the rhythmic turn-taking patterns in human conversation. In particular, Jaffe and Feldstein (1970) studied the mean duration of pauses, switching pauses (when a different speaker takes the floor), simultaneous speech, and (singlespeaker) vocalizations in recorded dyadic conversations. Based on their observation that these durations follow exponential distributions, they proposed first-order Markov models to capture the alternation of speech and silence in the dialog. Their initial model had four states: only participant A is speaking; only participant B is speaking; both participants are speaking, and neither participant is speaking. However, such a model fails to distinguish switching pauses from A to B from switching pauses from B to A. Based on this observation, they extend their model to a six-state model which they found to better fit their data than the four-state model. Around the same time, Brady (1969) developed a very similar six-state model. He trained the parameters on a recorded conversation and compared the generated conversations to the original real one along several dimensions (pause and speech segment durations, overlaps, etc), finding that his model generally produced a good fit of the data \cite {raux2009finite}.

\section{Conclusion}
This chapter presented the key fundamental concepts that are related to the creation of our chatbot. The next chapter will explain each component of the chatbot architecture apart.
\chapter{Chatbot Architecture}
\minitoc
\newpage
\section{Introduction}
This chapter will be devoted to the presentation of a new architecture that
integrates an Ontology, a Finite State Machine, a trained Chatbot and several other
modules that contributed to the construction of our chatbot.

\section{Architecture Components}
\begin{figure}[!ht] 
\includegraphics[height=8cm, width=13cm]{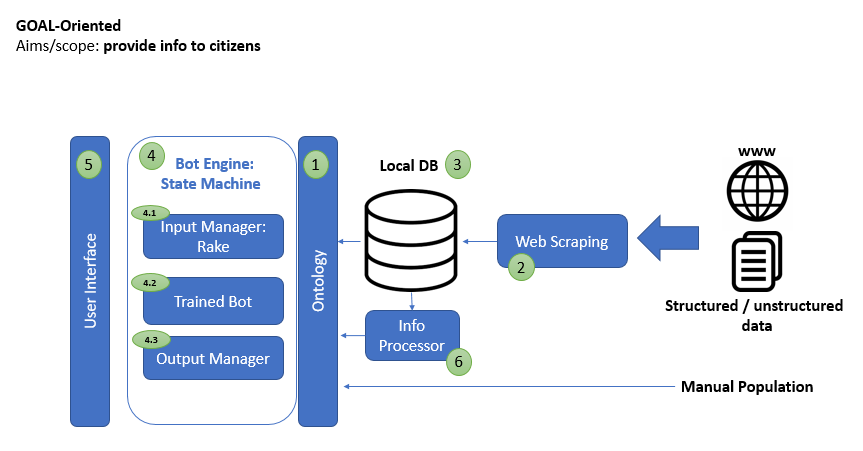}
\caption{Reference Architecture} 
\label{archi}
\end{figure}

Following this architecture in Figure\ref{archi}, we are going to implement an intelligent chatbot
support system for the COVID-19 virus, that will provide an effective response to
user’s queries. Our conversational system is a mixed system, the response
provided to the user can be generated either by information retrieved from the
ontology or by a trained bot.

As for the programming language of our chatbot we decided for Python. 

\newpage
\subsection{Ontology}
\begin{figure}[!ht]
\centering
\includegraphics[scale=0.7]{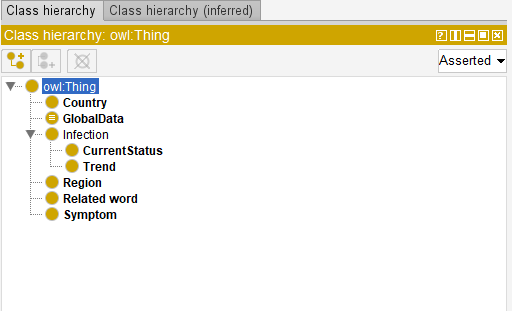}
\caption{Class Hierarchy of CVIO}
\label{cvio}
\end{figure}

We offer Corona Virus Infection Ontology (CVIO) which covers concepts related to COVID-19 virus such as countries that have COVID-19 infection, global data, and symptoms. \newline
To build such an ontology we used existing biomedical ontologies, we started by analyzing Human Disease Ontology (DOID) which provides disease information and Symptom Ontology (SYMP) which provides different symptoms of a certain type. They are published in BioPortal, one of the exceptional references, the world's largest online ontology repository hosting more than 450 biomedical ontologies and more than 6 million classes that define the terms  \cite{DBLP:journals/jbi/OchsHZGPHM16} . 
Indeed, biomedical ontologies are becoming increasingly important in order to provide an effective representation of knowledge to support the diagnostic process.
The next step is to define a global ontology that contain both disease and symptoms. Then we added other concepts like global data and the current state of contamination. \newline
The ontology engineering process spans different stages, below we will explain these stages one by one.\newline
\begin{enumerate}
\item \textbf{Knowledge acquisition.} Data is collected via different sources from the web through web scraping and articles related to coronavirus infection such as  \cite{10.1001/jama.2020.1585} from which we extract the main symptoms.The following table presents the symptoms in decreasing order of the percentage of appearance for the patients (there are others but with small percentage).
\begin{center}
\begin{table}[!ht]
 \centering
\begin{tabular}{|c|c|}
 \hline
 Symptom & Percentage \\
\hline
Fever & 98,6\%  \\
\hline
Fatigue & 69,6 \% \\
\hline
Dry cough & 59,4 \% \\
\hline
Myalgia  & 34,8 \% \\
\hline
Dyspnea  & 31,2 \% \\
 \hline

\end{tabular}
\caption{The main symptoms of COVID-19}
\end{table}
\end{center}
\item \textbf{Competency Questions (CQs.)}  After data acquisition, the first step towards developing the ontology is to define the vital questions that an ontology must answer. In the proposed project, for example, the competency questions can be: “Does Fever is a COVID-19 symptom?”, “Which countries have COVID-19 infection?'' etc. 
\item \textbf{Concepts. } Next step, is to define concepts, which concepts are necessary to answer the previous predefined questions. A concept is the basic building of an ontology, which can be designed as a set of individuals.  
\item \textbf{Properties.} The last step in the ontology engineering process is to define the properties, properties of the concepts, properties of individuals, and properties between individuals.
\end{enumerate}
The hierarchy of our ontology is presented in Figure\ref{cvio}.
\subsubsection{Classes::}
\begin{itemize}
\item \textbf{Country.} a concept that brings together countries that have COVID-19 infection;
\item \textbf{Global data.} a concept that represents the current status of the world; 
\item \textbf{Infection.} \begin{itemize}
\item \textbf{Current status.} a concept that represents the basic information about the state of the contamination of each country;
 \item \textbf{Trend.} a concept that represents trend for Covid-19 contamination. In our case, we manage a weekly trend which can be "weekly trend: increasing $/$ decreasing";
\end{itemize}
\item \textbf{Region.} a concept that contains the individual "world" which has a current status but it is not a country. Region is disjoined with the Country;
 \item \textbf{Symptom.} a concept that contains symptoms extracted from \cite{10.1001/jama.2020.1585} ;
\item \textbf{RelatedWord.} This class contains words related to COVID-19, it helps to manage the dialogue. When we find one of these words in the user's query, we know that the user is requesting general information regarding COVID-19 infection worldwide.
\end{itemize}

\subsubsection{Properties ::}
\begin{itemize}
\item \textbf{Data property.} is applied to each individual in current status concept. \textbf{Cases} (represent the total number of COVID-19 infected person), \textbf{ Recovered} (represent the number of recovered people), \textbf{Deaths} (represent the number of deaths) and \textbf{Retrieved} (represent the date on which this information are retrieved). \textbf{The domain} of the properties is current status concept or even Infection.

\item \textbf{Object property.} is applied between two individuals. In our ontology we defined two object properties \textbf{Region} and \textbf{Country}. For example: Tunisia\_Status Country Tunisia.
\item \textbf{Annotation Property.} is a property and can be applied to every thing. In our ontology we defined \textbf{Has Synonym} used for symptom individuals, \textbf{Creator} and \textbf{Creation date} for the ontology and \textbf{Data Source} and \textbf{ Data Publisher} for information in current Status individuals.

\newpage
\end{itemize}
\begin{figure}[!ht]
\centering
\includegraphics[scale=0.7]{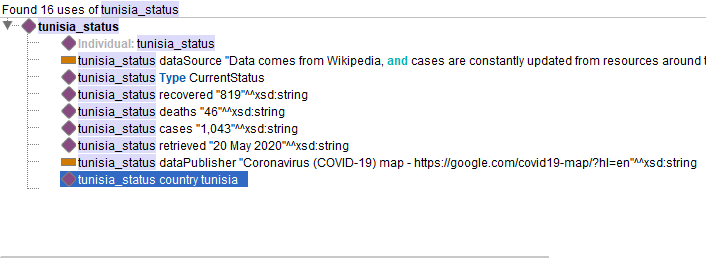}
\caption{Example of an individual}
\label{ex}
\end{figure}
Figure \ref{ex} shows an example of a current status individual \textbf{tunisia\_status} which has the object property (Country), the data properties (Cases, Recovered, Deaths, Retrieved) and Annotation properties (Data Source, Data Publisher).
\subsection{Web scaping}
\begin{figure}[!ht]
\centering
\includegraphics[scale=0.4]{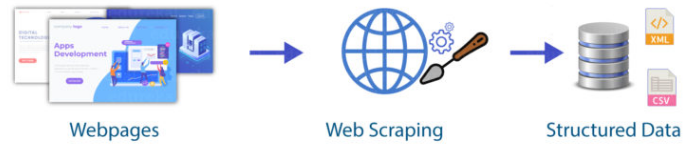}
\caption{Web scraping technique}
\label{fihkh1}
\end{figure}
The ontology is populated with data from a web site, which is continuously updated, in this case, we need to apply web scraping techniques. Web scraping is defined as: "a tool for turning the unstructured data on the web into machine-readable, structured data which is ready for analysis"\footnote{ https://www.freecodecamp.org/news/better-web-scraping-in-python-with-selenium-beautiful-soup-and-pandas-d6390592e251/}. \newline
The source of the data must be a very reputable website to guarantee the correctness of information populated to the ontology, also if the page is not stable enough the crawler won’t work anymore. For this we choose to scrap data from google but crawling was not allowed, so we scraped data from wiki site\footnote{\url{https://en.wikipedia.org/wiki/COVID-19\_pandemic}} in which coronavirus table is referred by google site\footnote{\url{https://google.com/covid19-map/?hl=en}}.
\begin{figure}[!ht]
\centering
\includegraphics[scale=0.7]{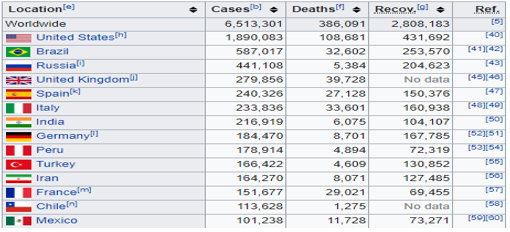}
\caption{Current status of different country}
\label{tab}
\end{figure}
The website we are going to use contains a table in which there is the current status of every country that has covid-19 infection shown in the Figure\ref{tab}. To do so, we used libraries such as \textbf{ BeautifulSoup} which is a package for parsing HTML and XML documents. It creates parse trees that are helpful to extract the data easily and \textbf{Selenium} which is a web testing library. It is used to automate browser activities. Finally, data retrieved are stored in a local database and arranged by retrieval date.
\subsection{Database}
Added to the ontology which stores the most recent retrieved data, we store the data in a local database. The use of a DBMS (Database Management System), that is requested anytime the amount of the information you deal with becomes relevant. Using a file (in XML format in this case) is unreliable and not practical at all. In concrete, a database is requested because:

PERFORMANCE. The size of the ontology will increase quite soon if we store all the data retrieved on it. So the idea is to only keep the most recent status (e.g. today’s data concerning infections) in the ontology and the computed info (component 6 in the architecture), i.e. “Weekly trend: decreasing”, “Weekly trend: increasing”. Raw data will  ONLY be stored in the database. 

COMPUTATIONS. Performing computations on semantic data, even with the support of SPARQL (the equivalent of SQL for semantic data), can be a nightmare once the algorithm is not obvious, while it is relatively simple on raw data stored in a common format.

DATA CONSISTENCY. The retrieval components and the Chatbot have different status. Updating the ontology means to update the reasoner status too. It is advisable to keep the database in the middle to assure flexibility and independent behaviors.

\subsection{State Machine}

\begin{figure}[!ht]
\includegraphics[width=14cm, height=12cm]{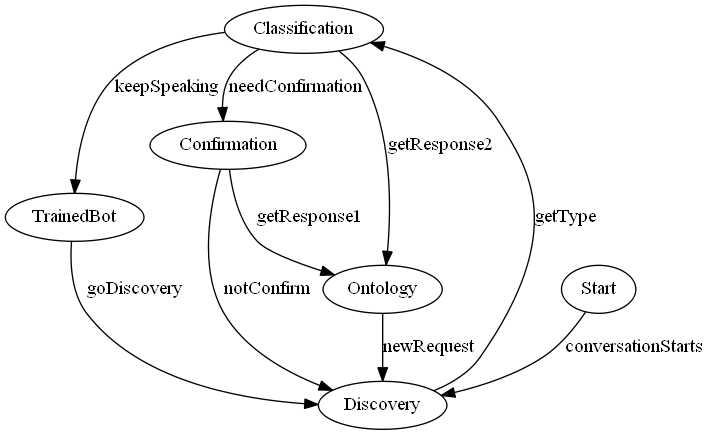}
\caption{State Machine Structure}
\label{fsm}
\end{figure}

In this section, we are going to describe the engine of our COVID Assistant which is a state machine.
In order to model dialogues, we made 6 states and defined 9 transitions using a python library. When our system is initialized, it enters ‘Start’ state. The structure of the state machine is presented in the Figure\ref{fsm}.
\newpage
\begin{itemize}
\item \textbf{Start.} It represents the initial state, in which the chatbot will present itself in a fun and informative way. The user has the right to know whether he is speaking with a real person or with a bot. For this, the welcome message must clearly indicate that the customer is chatting with a chatbot. In addition, the chatbot gets the name of the user.
\item \textbf{Discovery.} Discovery state is the state in which the user types the input/request. Then, a keyword extractor is applied (RAKE) in our case. It extracts the keywords from the user input and sorts them by order of importance.
\item \textbf{Classification.} We go now to the classification state where we iterate the list of keywords, and we try to understand the user request.
In this state, we manage four classes. First, when the user request contains the name of a specific country. In this case, we understand that the user wants to get information about the current state of this country, Then we pass to the confirmation state. Second, the request may contain the name of a symptom of covid-19, so we can provide him with a list of the most common symptoms. 
The third classification is that the user would like to know about global data something like the current state of the world in this case, he just has to type any word related to COVID-19.
The last type of classification is that we don't know what the user is talking about, it can be a welcome message, a goodbye or a general conversation, in this case, we forward the request to the TrainedBot state. \newline We store the keyword, after classification, in our context variable except the last type so we can respond appropriately. 
\item \textbf{TrainedBot.} The purpose of the Trainedbot state is to try to get the user to speak when we don't understand him. Hopefully, we can find a keyword in the next entry. In this case, an automatic response is generated.
\item \textbf{Confirmation.} If the state machine is in the Confirmation state, this means that we have found a country name, we send a message asking for user confirmation. If the answer is "no", we return to the Discovery state, otherwise we go to the Ontology state.   
\item \textbf{Ontology.} The ontology state is the final state in which, depending on the value we have in the context variable. We extract the appropriate information from the ontology and, finally, the user obtains the response to his request. 
\end{itemize}
Now we will describe the meaning of each transition apart: 
\begin{itemize}
\item \textbf{Conversation Starts. }This transition executed once the system obtains the user's name, at that point the conversation will start.
\item \textbf{Get Type.} After getting input from the user "Get Type" transition will be executed in order to understand the user's request.
 \item \textbf{Get Response2.} When the system is in "Classification State", it has three possibilities, one of them is "Get Response2" which means the user wants information either about symptoms or about COVID-19 general information.
 \item \textbf{Need Confirmation.} The second possibility, is when the user request contains one or more country names, so the "Need Confirmation" transition is executed to pass the system into the "Confirmation State". 
\item \textbf{keep Speaking.} Otherwise, the "Keep Speaking" transition is executed, this transition exchanges the current status of the system and executes the "Trained Bot" state.
\item \textbf{New Request.} This transition is executed when the chatbot has already answered the user and let him type a new request.
\item \textbf{Get Response1.} If the system gets a positive confirmation result it means that the country name it has is correct. So, this transition exchanges the current state of the system to the "Ontology" state. 
\item \textbf{Not Confirm.} This transition means that the system gets negative confirmation, so this transition will go to the "Discovery" state where the user can type another request. 
\item \textbf{Go Discovery.} This transition has a similar functionality as "New Request". The source state of the transition is the only difference between the two.
\end{itemize}
The state machine module is developed with a state machine library, access to the ontology data is done with owlready2. 
\newpage 
\subsection{Keyword Extraction}
In this section, we will explain how we extract the keywords from our input. Keyword extraction (also known as keyword detection) is a text analysis technique that is concerned with the automatic extraction of a set of terms or phrases within unstructured text. This includes emails, social media posts, chat conversations and any other types of data that are not organized in any predefined way. \\
This task is commonly used for information retrieval, text mining, summarizing, and other text analysis tasks. The conventional method for extracting keywords is the TD-IDF measure, which targets words that frequently appear in one document in a corpus but rarely appear in the rest of the corpus. \\
There are different techniques you can use for automated keyword extraction: statistical approaches, Linguistic Approaches, Graph-based Approaches, Machine Learning Approaches. \\
We choose a statistical approach, Rapid Automatic Keyword Extraction (RAKE) is a well-known keyword extraction method that uses a list of stopwords and phrase delimiters to detect the most relevant words or phrases in a text. 
RAKE was developed by Rose, Engel and Cramer  \cite{rose2010automatic} as an alternative to corpus-based keyword extraction methods. Unlike traditional methods that compare word frequencies, RAKE operates on the premise that keywords and key phrases do not overlap with stopwords, and that the more non-stop words are found adjacent to each other, the higher the probability that it is a keyphrase  \cite{leung2016evaluating}.
\begin{itemize}
\item \textbf{Candidate keywords.} To produce the list of candidate keywords/phrases, the first thing the method does is splitting the text into a list of words and remove stopwords from that list. This returns a list of what is known as content words. Candidates do not include phrase delimiters or stopwords.
\item \textbf{Co-occurrence graph. }Once the candidate keywords are chosen, a co-occurrence graph is built to identify the frequency that words are associated together in those phrases. 
\item \textbf{keyword scores. }After that graph is built, words are given a score. That score is calculated for each individual word as the degree (number of times it appears + number of additional words it appears with) of a word divided by its frequency (number of times it appears), which weights towards longer phrases. The expressions are also given a score, which is computed as the sum of the individual scores of words.
\item \textbf{Adjoining keywords. }If two keywords or phrases appear together in the same order more than twice, a new key phrase is created, regardless of how many stop words the key phrase contains in the original text. The score of this key phrase is calculated as the score of a single key phrase.
\item \textbf{Extracted keywords. }The top T keywords are then extracted from the content, where T, According to the original paper  \cite{rose2010automatic} is one-third of the number of words in the graph
\end{itemize}
\subsection{Trained Chatbot}
For general conversation management, we integrated a trained chatbot component in our architecture. This component implements Chatterbot. 

\begin{wrapfigure}[]{l}{6cm}
\includegraphics[width=0.5\textwidth]{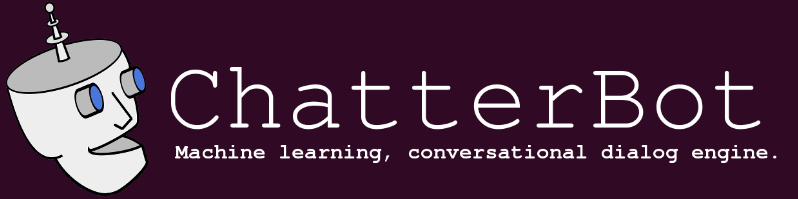}
\caption{ChatterBot}
\label{chatter}
\end{wrapfigure}
ChatterBot is a machine-learning based conversational dialog engine build in Python which makes it possible to generate responses based on collections of known conversations. The language independent design of ChatterBot allows it to be trained to speak any language. It uses a selection of machine learning algorithms to produce different types of responses. 

\begin{figure}[!ht]
\centering
\includegraphics[width=9cm]{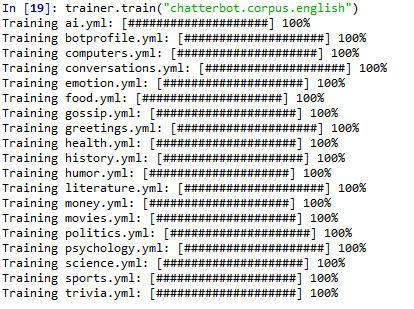}
\caption{Training Packages}
\label{training}
\end{figure}

ChatterBot comes with a data utility module that can be used to train chatbots. At the moment there is training data for over a dozen languages in this module. These modules are used to quickly train ChatterBot to respond to various inputs. Thus, it is trained with different packages such as greetings, gossip, history, etc. as shown in Figure\ref{training}.

\newpage 
\subsection{User Interface}
The last component in our architecture is the user interface (UI), which a very important one and has a great impact on the usability of the chatbot.

Customers will likely abandon your chatbot if it can’t keep up with them or is too frustrating to use. Putting careful thought into your chatbot’s user interface is the first step to avoiding this.

\section{Conclusion}

In this third chapter, we discussed the architecture of Covid-19 Assistant and in the next chapter, we will describe the conceptual study and the implementation details.  
\chapter{Conceptual study and Implementation}
\minitoc
\newpage
\section{Introduction}

After the analysis and specification of our system, we dedicate this chapter to conceptual study. This design phase has great importance in our development cycle and whose objective is to create a robust, well structured, and scalable system.
The second part of this chapter will be devoted to the implementation where the technical specification will be discussed.

\section{Conceptual study}

The modeling of our project was based on the Unified Modeling Language (UML).
\subsection{Static view}
In this part, we start by designing the general use case diagram. This use case diagram (UCD) is a UML diagram used to express user needs. In other words, it allows us to give a global vision of the functioning of the chatbot.
\begin{figure}[!ht]
\centering
\includegraphics[width=13cm, height=9cm]{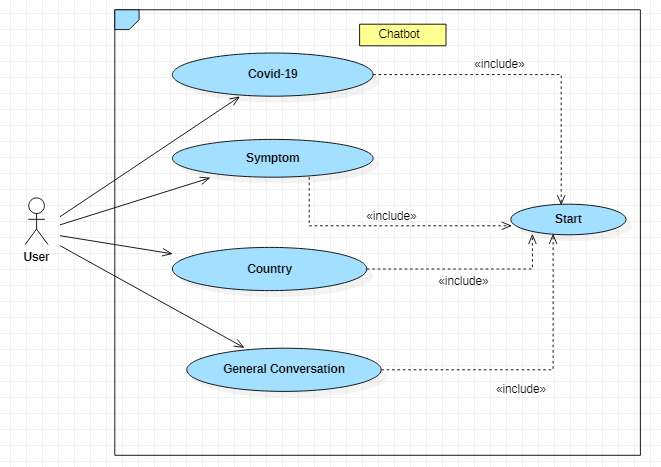}
\caption{General use case diagram}
\label{ucd}
\end{figure}

In the case of our chatbot, a user can be, an internet user, who will ask the chatbot for information, namely the state of contamination of COVID-19 in the world Figure \ref{ucd} illustrates the general use case diagram that allows us to see the functionality provided to our actor.

\begin{center}
\begin{table}[!ht]
\centering
\begin{tabularx}{13cm}{|c|X|}
\hline
\cellcolor{lightgray}Action & \cellcolor{lightgray} Description  \\
\hline
\cellcolor{lightgray} Start & This service represents the initial state of our chatbot. It lets the user know that he is talking to a chatbot while giving him the illusion that he is talking to a human assistant.

Then the user must enter his name in the chatbot to go to the next state.\\
\hline
\cellcolor{lightgray} Covid-19  & Once the knowledge step is done:
the user can request to know what is Corona Virus? or the current state of contamination of Corona Virus infection in the world. \\
 \hline
 \cellcolor{lightgray} Symptom  & 
To know the different symptoms of COVID-19: 
 
 the user just needs to type in the request the word symptom or any symptom of Coronavirus. \\
 \hline
 \cellcolor{lightgray} Country  & To know the contamination status of a specific country:

 the user has to enter the name of the country or the name of any city it contains. \\
 \hline
\cellcolor{lightgray} General Conversation &  This state will be visited if the user enters a request that does not contain any keyword that the chatbot might know.
 
 So the trained chatbot will respond. \\
\hline
\end{tabularx}
\caption{General Use Case Diagram Table}
\end{table}
\end{center}

\subsection{Dynamic view}
The dynamic axis in our design will be represented by sequence diagrams. These diagrams are the graphic representation of the interactions between the actors and the system in chronological order. A sequence diagram describes a scenario of a use case. In the following, we will detail the different diagrams:
\newpage
\subsubsection{Sequence diagram "Get started"}
This sequence diagram represents the procedure of getting started to talk with the chatbot. Once the user opens the chat application the chatbot begins by sending welcome message and introducing himself. Then he asks the user for his name, if the user answers it in a good way it retrieves the name, else, it re-send an example of the way he must answer with. To represent this we used a loop and alternative (alt) fragments.
\begin{figure}[!ht]
\centering
\includegraphics[width=13cm, height=9cm]{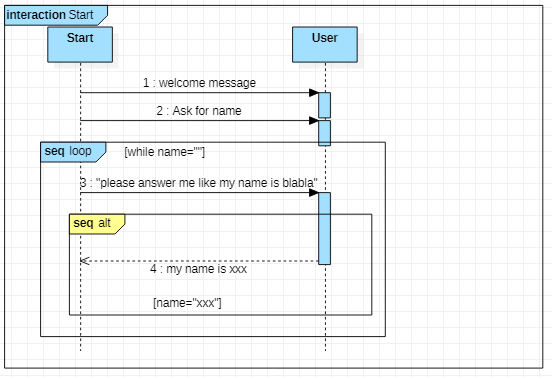}
\caption{Management of name retrieval}
\end{figure}

\subsubsection{Sequence diagram: "Managing a request for a country name (version 1)"}
To search for information about a specific country the user has to type the name of a country or a city that it contains like in the example. The user didn't type Tunisia but he types the capital name Tunis. The keyword, in this case, is "Tunis", then in the classification state, the system will associate "Tunisia" to "Tunis" automatically. Once a country name is found  the chatbot will ask the user for the confirmation. To represent it we use a loop and an alternative fragment, at this stage the user has to type "yes" or "no" if he types anything else, the chatbot re-send to him the confirmation message.
\begin{figure}[!ht]
\centering
\includegraphics[width=13cm, height=13cm]{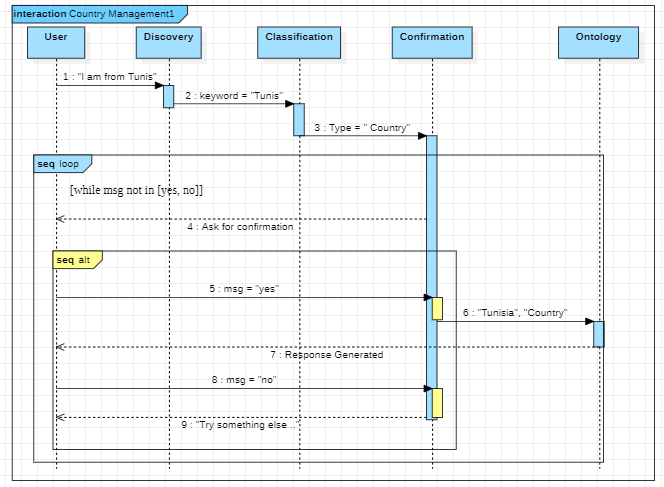}
\caption{Management of a country request (version 1)}
\end{figure}

\subsubsection{Sequence diagram: "Managing a request for a country name (version 2)"}
In this second version of managing a request for the country information, the user types a message that contains more than one country name. In this case, after recognizing the country names in the classification state, the chatbot will send to the user the list of country names and the user has to choose one of them by typing the number of the name in the list. Until a correct number is typed the system will still in the loop and alternative frame.  Once a correct number is entered the system pass to the ontology state and sends to the user the appropriate answer. 
\begin{figure}[!ht]
\includegraphics[width=13cm, height=13cm]{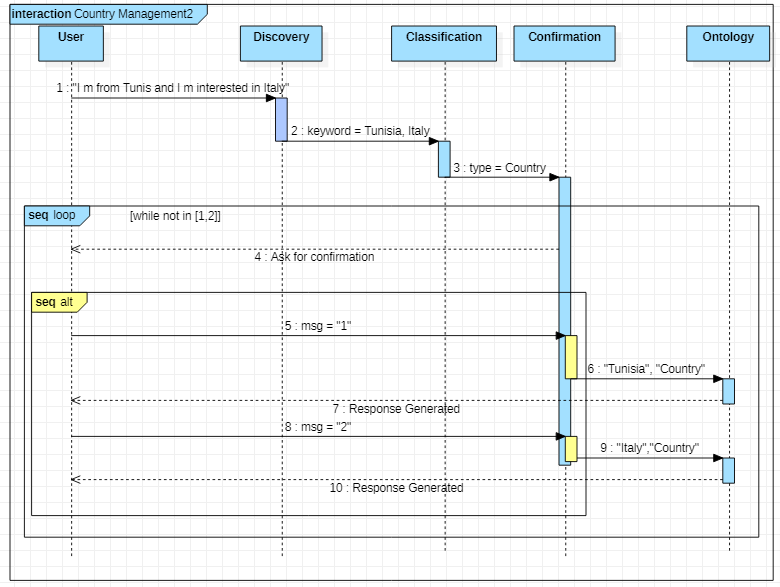}
\caption{Management of a country request (version 2)}
\end{figure}

\newpage
\subsubsection{Sequence diagram: "Managing a request of covid-19 general information"}

\begin{figure}[!ht]
\centering
\includegraphics[width=13cm, height=6cm]{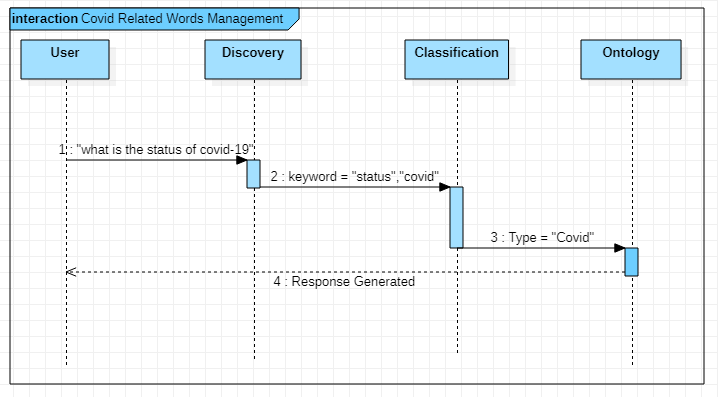}
\caption{Management of COVID-19 request}
\end{figure}
\subsubsection{Sequence diagram: "Managing a request related to COVID-19 symptoms"}
\begin{figure}[!ht]
\centering
\includegraphics[width=13cm, height=6cm]{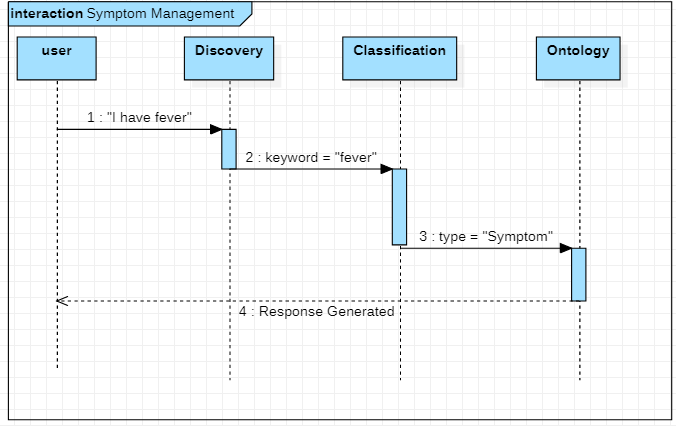}
\caption{Management of symptom request}
\end{figure}
In the Sequence diagram "Managing a request related to COVID-19 symptoms" and Sequence diagram "Managing a request of COVID-19 general information" the same procedure is required. The first step, of course after passing the start state, the user types a message that contains a keyword related either to COVID-19 in general or a name of a symptom. The discovery state will extract keywords then the classification state will recognize the type of this keywords. Finally, the system passes the type detected to the ontology state which provides the appropriate response to the user.    
\subsubsection{Sequence diagram: "Managing a general conversation"}
\begin{figure}[!ht]
\centering
\includegraphics[width=13cm, height=9cm]{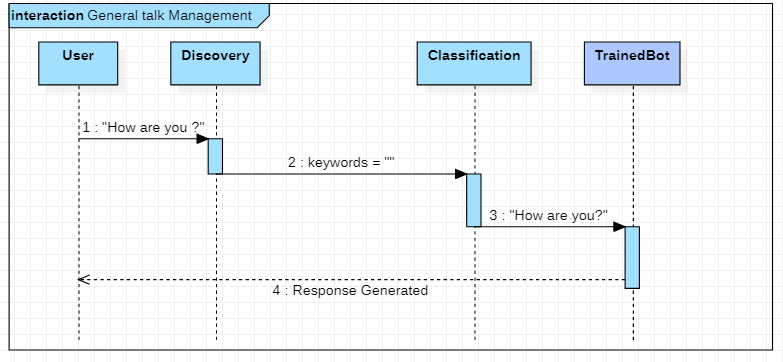}
\caption{Management of general talk request}
\end{figure}
Two types of messages lead the conversation to do this interaction, either the user types a message like in the example shown that do not contain any keyword or it contains keywords that the system can't classify it.  So, it passes the message to a trained bot that is trained to answer the general conversations.
\newpage
\section{Implementation}
\subsection{Python }
\begin{wrapfigure}[]{l}{4cm}
    \includegraphics[width=0.3\textwidth]{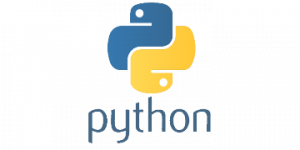}
  \caption{Python}
\end{wrapfigure}
Python\footnote{\url{https://www.python.org}} is an interpreted, multi-paradigm, multi-platform programming language designed by Guido van Rossum in February 20, 1991. It supports structured, functional, and object-oriented imperative programming. It has strong dynamic typing, automatic garbage collection memory management, and exception handling, making it similar to Perl, Ruby, Scheme, Smalltalk, and Tcl. In our chatbot, we used python 3.5 version. We used anaconda\footnote {\url{https://www.anaconda.com}} as it is a free and open-source distribution, applied to perform data science and machine learning applications, with over 20 million users worldwide. 
\subsection{Protégé} 
\begin{wrapfigure}[]{l}{4cm}
 \includegraphics[width=0.3\textwidth]{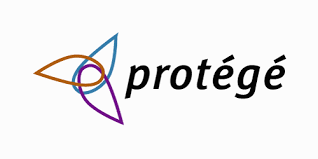}
\caption{Protégé}
\end{wrapfigure}
Protégé\footnote{\url{https://protege.stanford.edu}} plug-in architecture can be adapted to build both simple and complex ontology-based applications. Developers can integrate the output of Protégé with rule systems or other problem solvers to construct a wide range of intelligent systems. To visualize our ontology we used OWLViz which Enables class hierarchies in an OWL ontology to be viewed and incrementally navigated, allowing comparison of the asserted class hierarchy and the inferred class hierarchy.

\newpage
\subsection{MySQL/ MySQL Workbench }
\begin{wrapfigure}[]{l}{4cm}
 \includegraphics[width=0.3\textwidth]{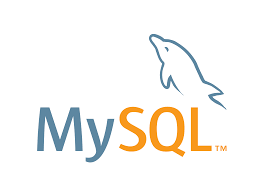}
\caption{MySQL}
\end{wrapfigure}
MySQL\footnote{\url{https://www.mysql.com}} is the world's most popular open-source database. Despite its powerful features, MySQL is simple to set up and easy to use.
MySQL Workbench\footnote{\url{https://www.mysql.com/fr/products/workbench/}} is a unified visual tool for database architects, developers, and DBAs.
MySQL Workbench enables a DBA, developer, or data architect to visually design, model, generate, and manage databases. It includes everything a data modeler needs for creating complex ER models, forward and reverse engineering, and also delivers key features for performing difficult change management and documentation tasks that normally require much time and effort.

\subsection{StarUML}
\begin{wrapfigure}[]{l}{4cm}
 \includegraphics[width=0.2\textwidth]{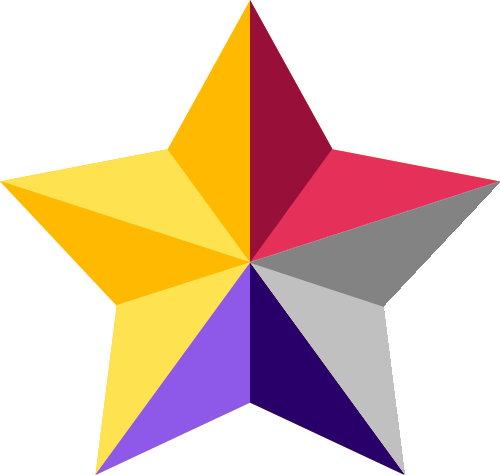}
\caption{StartUml}
\end{wrapfigure}
StarUML\footnote{\url{http://staruml.io}} is a sophisticated software modeler aimed to support agile and concise modeling. It is a UML modeling software, which has been released as open source by its publisher, at the end of its commercial exploitation (which obviously continues\ldots), under a modified license of GNU GPL.
Today the version StarUML V3 exists only in proprietary license. StarUML supports most of the diagrams specified in the UML 2.0 standard. StarUML is written in Delphi, and it depends on proprietary which is non open-source Delphi components.
\newpage
\subsection{Uppaal}
\begin{wrapfigure}[]{l}{5cm}
\includegraphics[width=0.4\textwidth]{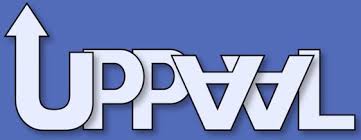}
\caption{Uppaal}
\end{wrapfigure}
Uppaal is a toolbox for verification of real-time systems jointly developed by Uppsala University and Aalborg University. It has been applied successfully in case studies ranging from communication protocols to multimedia applications. The tool is designed to verify systems that can be modelled as networks of timed automata extended with integer variables, structured data types, user defined functions, and channel synchronisation.

\subsection{Python libraries}

\begin{center}
\begin{table}[!ht]
\centering
\begin{tabularx}{13cm}{|c|X|}
\hline
\cellcolor{lightgray}Library & \cellcolor{lightgray} Description \\
\hline
\cellcolor{lightgray} pycountry  &  This python library is used in the Classification state. It is capable of returning a country name from a city name. \\
\hline
\cellcolor{lightgray} geocoder  & This library helped us is the creation of the data visualisation (Covid-19 map). It provides the location of a given country\\
\hline
\cellcolor{lightgray} tkinter & The User Interface component is created with tkinter library. \\
 \hline
 \cellcolor{lightgray} chatterbot  & We used chatterBot because of it's ability to respond to general conversations. \\
  \hline
 \cellcolor{lightgray} rake-nltk & Rake is used in Discovery state, in which our chatbot tries to understant the request of the user by extracting the important keywords. \\
  \hline
  \cellcolor{lightgray} state-machine & This library is the engine of our chatbot. The state machine describes the workflow of the system. \\
 \hline
 \cellcolor{lightgray} graphviz  & This library is used to visualize the state machine. \\
 \hline
 \cellcolor{lightgray} owlready2  & Owlready2 makes the management of the ontology simple. It enables the creation, the removal of individuals.  \\
 \hline
 \cellcolor{lightgray} beautiful soup & This library makes HTML code easy to be scrapped. \\
 \hline
 \cellcolor{lightgray} selinum  & This library is used to automate web browser interaction from Python. \\
 \hline
  \cellcolor{lightgray} folum & This library is used to visualize the data on the Covid-19 map.\\
 \hline
 
\end{tabularx}
\caption{List of Python libraries}
\end{table}
\end{center}

\section{Conclusion}
In this chapter, we have described our design as well as our working environment. The next chapter will be devoted to the experiment and a validation step.
\chapter{Result}
\minitoc
\newpage
\section{Introduction}
This chapter of experiments will allow us to see the execution of our chatbot which proved its efficiency, and capability of maintaining a conversation while generating good quality responses to the user. The second part of this chapter will be devoted to a validation process in which we validated the ontology and the state machine components.

\begin{figure}[h!]
\centering
\includegraphics[width=7cm]{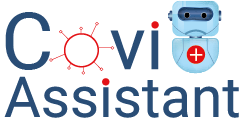}
\caption{Covid Assistant}
\end{figure}

\section{Execution}
In this part we will present some examples of conversation scenarios that can be modeled between the Chatbot and the user. These scenarios are proved to be compatible with the conceptualization defined above. The figures bellow show how our Chatbot react. Its response shows its capability of creating a conversation having a human aspect.

\begin {figure}[htbp]
  \hbox{ 
     \includegraphics[width=6cm]{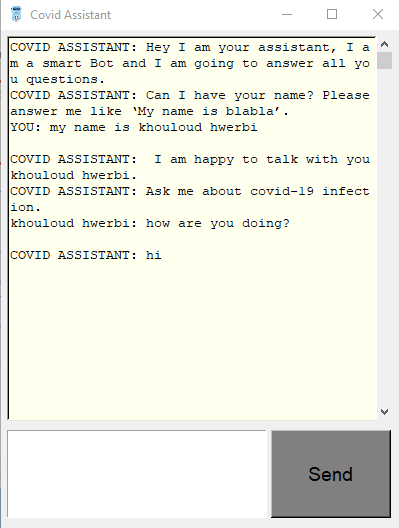}
     \hspace*{1cm}  
     \includegraphics[width=6cm]{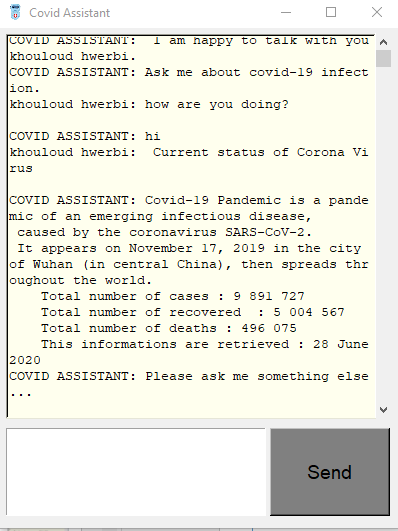}
  }
  \caption{Example 1 and 2}
  \label{ex12}
\end {figure}

Use case shown in Figure \ref{ex12} corresponds to \textbf{sequence diagram: "Get started"} in which the chatbot manage the name retrieval task and the \textbf{Sequence diagram: "Managing a request of covid-19 general information"} respectively.
\newpage

As the user asks about covid-19 general information the chatbot provides to him: what is covid-19 pandemic?, Numbers that represent the current status of contamination in the world and a Covid-19 Map. The map will be explained in the next section.

\subsubsection{Data Vizualisation}
Data visualization techniques are applied to provide a visual represntation of the data we have. Thanks to libraries like folium which is used here to create a COVID-19 map of contamination distributions. Also we used pycountry and geocoder libraries to create this attractive and informative map.
\newpage
\begin{figure}[h!]
\centering
\includegraphics[width=12cm]{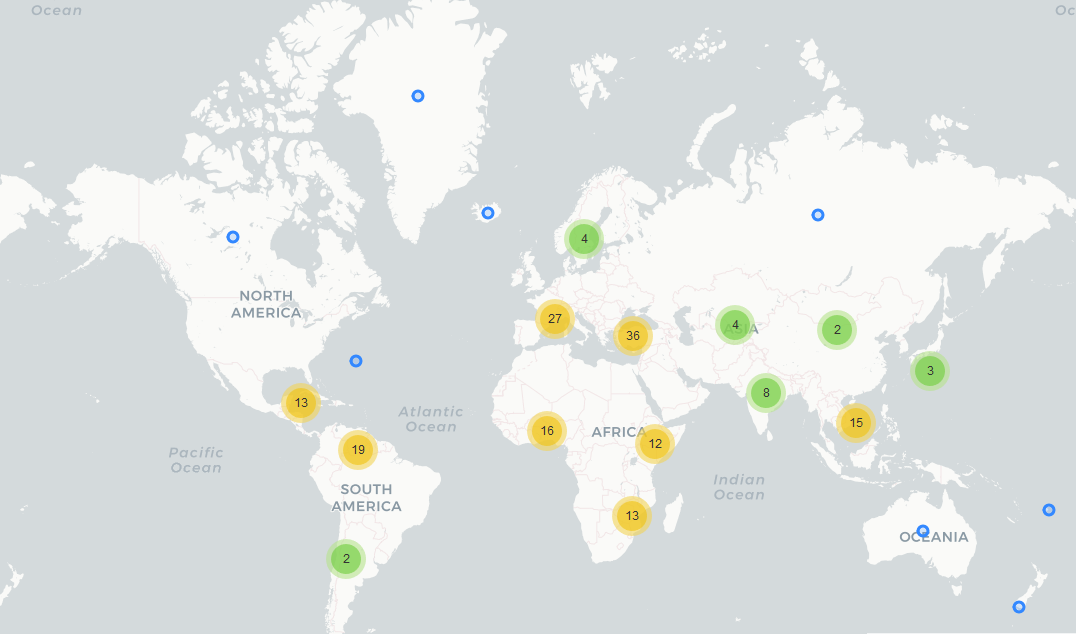}
\caption{Covid-19 Map}
\label{map}
\end{figure}
Data scraped from the web can be viewed as an interactive visual representation, so we create an attractive and informative map of certain numbers that shows the Covid-19 distributions around the world. When the user asks about global data, the response of our Covid-19 assistant will be like in a specific country use case, a message that contains: the total number of cases, the number of deaths, recovered people and an interactive map, which we called Covid-19 Map \ref{map}.

\begin{wrapfigure}[]{l}{6cm}
    \includegraphics[width=0.45\textwidth]{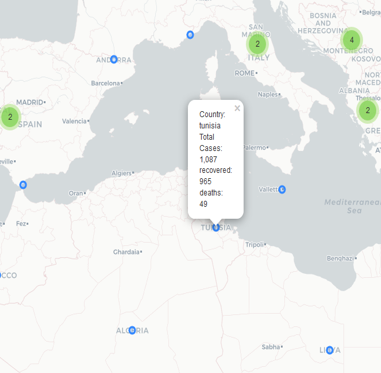}
  \caption{Current Status of Tunisia}
  \label{mapTun}
\end{wrapfigure}

This second map, Figure \ref{mapTun}, represents an example of data for a specific country, Tunisia, the user then can zoom in the principal map to see details about the current status of each country apart. Maps make the visualization of available data more effectively. It displays the geographical related data and presents the matching information on the map, this kind of information expression is clearer and  it will be eye-catching. We can visually see the distribution or proportion of data in each region.

\begin {figure}[htbp]
  \hbox{ 
     \includegraphics[width=6cm]{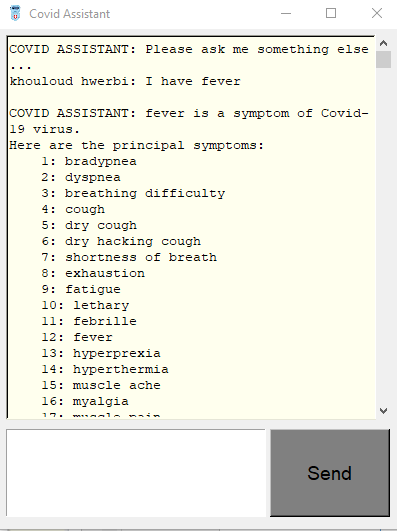}
     \hspace*{1cm}  
     \includegraphics[width=6cm]{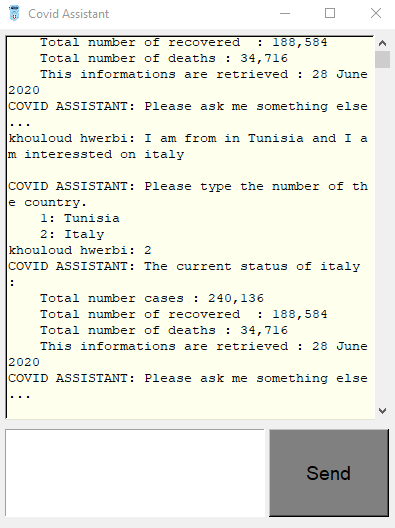}
  }
  \caption{Example 3 and 4}
  \label{ex34}
\end {figure}

\newpage
Now we illustrate in this example, Figure \ref{ex34} how the Chatbot answers a request of information concerning symptoms which is designed by \textbf{Sequence diagram: "Managing a request related to covid-19 symptoms"}. When the system extracts a word that matches with corona virus symptoms it responds by listing the principal symptoms that are stored in the ontology and their synonyms. Figure 5.5 represents also the scenario of \textbf{Sequence diagram: "Managing a request for a country name (version
2)"} is executed.

The next Figure \ref{ex56} represents scenarios explained in \textbf{Sequence diagram: "Managing a request for a country name (version 2)"} and \textbf{Sequence diagram: "Managing a general conversation"}.

\newpage
\begin {figure}[htbp]
  \hbox{ 
     \includegraphics[width=6cm]{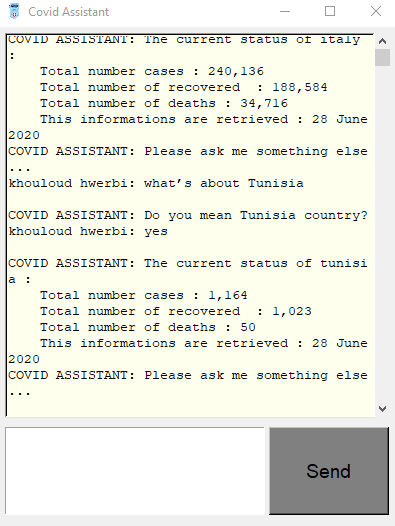}
     \hspace*{1cm} 
     \includegraphics[width=6cm]{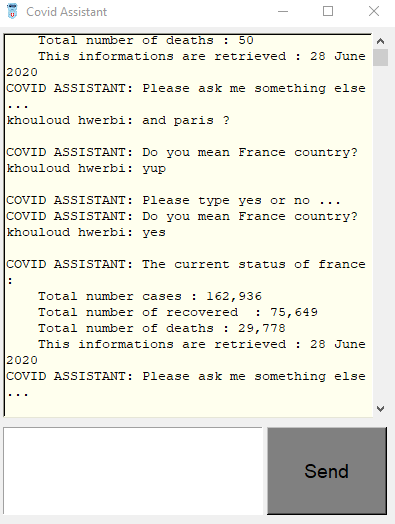}
  }
  \caption{Example 5 and 6}
  \label{ex56}
\end {figure}

\newpage
\section{Validation}
The objective is that the chatbot delivers answers to citizen's questions. This developed chatbot has reached this objective as seen in the example of executions in the first part. Now we will move on to validation two principal components of the chatbot architecture which are the ontology, our knowledge base, and the finite state machine, the engine of the chatbot.

\subsection{Ontology}
It should be noted that the validation of ontologies is a central question in knowledge engineering. It revolves around two complementary issues: \textit{i}) structural validation and \textit{ii}) semantic validation \cite {richard2015lovmi}.

\subsubsection{Structural validation}
\begin{itemize}
    \item Validation of consistency using PELLET \\ To validate our ontology we used PELLET reasoner. The choice of the PELLET reasoner to validate the consistency of our ontology was a natural one, given that we are using the Protégé 5.5.0 ontology editor in which it is integrated. PELLET thus allowed us to verify that our ontology did not contain contradictory classes to lead to an inconsistency state.

\end{itemize}
\subsubsection{Validation of semantics: Suitability for the modeled domain}
\begin{itemize}
    \item This stage brings into play communication aspects between actors from different areas of expertise: ontologists and specialists in the domain modeled in ontology.
    In our case the ontology is a functional component which organizes and integrates data in a way that is directly usable by the chatbot as an unique source. So, if our use cases are supported by the chatbot, then, the ontology is validated.

\end{itemize}

\newpage
\subsection{State machine}

To validate the engine of the chatbot we used a model checker. Model checker or property controller is a method of checking whether a finite state machine of a system satisfies a given specification or accuracy. 

In an attempt to solve such a problem algorithmically, both the finite state machine of the system and its specification are formulated in a precise mathematical language. For this purpose, we used the Uppaal model checker. Uppaal is a toolbox for validation (by graphical simulation) and verification (by automatic model verification) of real-time systems. 

\begin{figure}[h!]
\centering
\includegraphics[width=10cm, height=6cm]{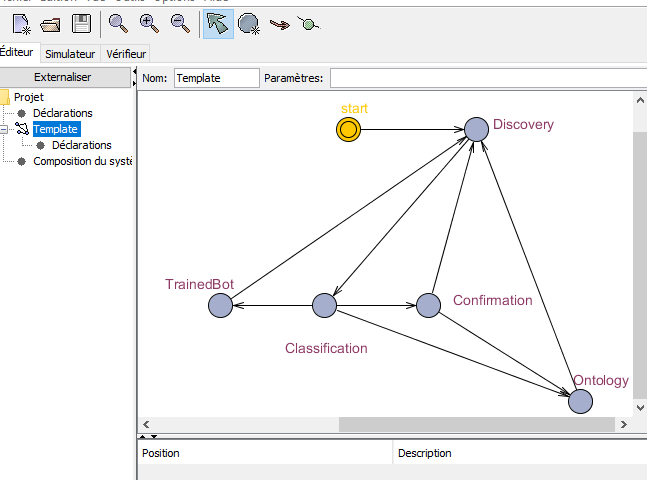}
\caption{Model desinged}
\label{uppaal}
\end{figure}

The requirement specification must be expressed in a formally well-defined and machine readable language. Uppaal uses a simplified version of CTL.
We first create our state machine as shown in the Figure \ref{uppaal}.  
we will start by demonstrating the sequential composition of the model and then demonstrate the parallel composition
\subsubsection{Sequential Composition:}

\begin{figure}[h!]
\centering
\includegraphics[width=10cm, height=6cm]{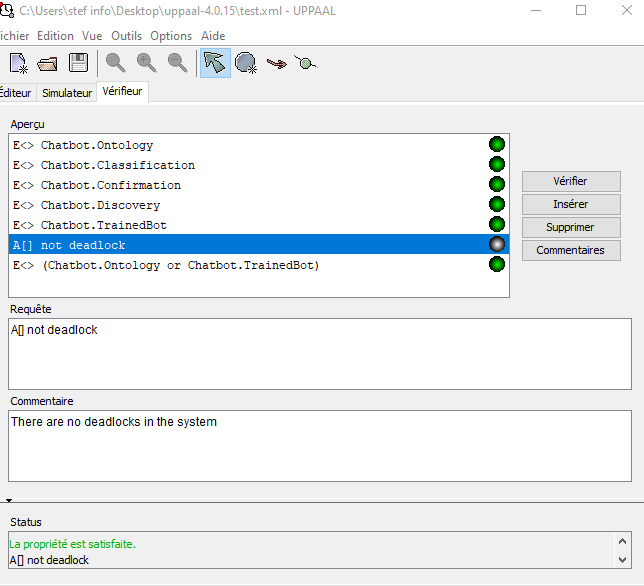}
\caption{Properties Verification}
\label{valid}
\end{figure}

For our model we are going to check reachability, safety, and liveness properties.
\begin{itemize}
    \item \textbf{Reachability Properties. }are the simplest form of properties. They ask whether a given state formula, $\varphi$, possibly can be satisfied. Does there exist a path starting at the initial state, such that $\varphi$ is eventually satisfied along that path.
    we use the formula:\textbf{ E$<>$ Chatbot.Ontology} to verify if the Ontology state is reachable. We did the same thing with all other states.
    \item \textbf{Safety Properties. }are on the form: “something bad will never happen”.  We used this formule 
    \textbf{ E$<>$ (Chatbot.Ontology or Chatbot.TrainedBot)} to grantee there should exist a maximal path such that a response is always generated to the user.

    \item \textbf{Liveness Properties. }are of the form: something will eventually happen, We check that the system is deadlock-free with the property\textbf{ A[] not deadlock}

\end{itemize}
All the properties previously discussed are shown in the Figure \ref{valid}.

\subsubsection{Parallel Composition:}
\begin{figure}[h!]
\centering
\includegraphics[width=10cm, height=6cm]{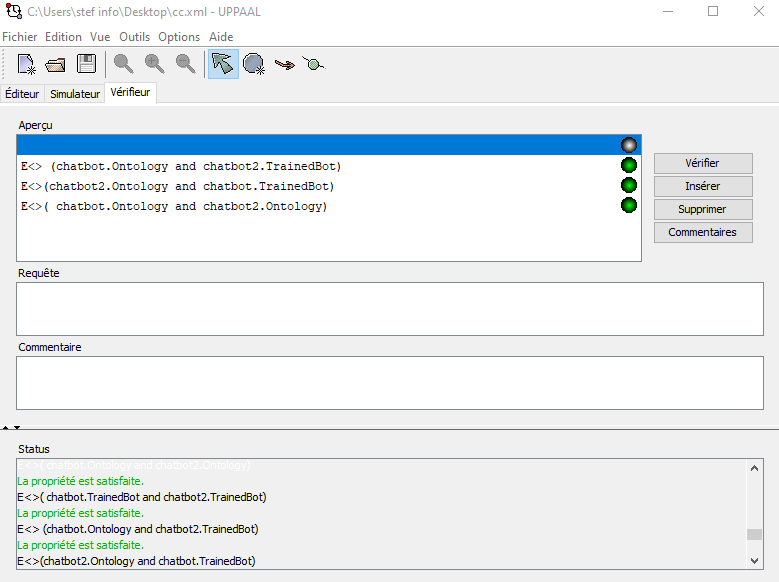}
\caption{Parallel verification}
\label{parallel ver}

\end{figure}

For checking the parallel aspect in our chatbot we have created two instances of the system: chatbot and chatbot2.  We will check that the user will always get a result either from the ontology or through the TrainedBot. 
To do this, the following three formulas are created 
\textit{i}) there is an execution of the system that allows us to put the first instance "chatbot" in the Ontology state and the second instance "chatbot2" in the TrainedBot state through the formula \textbf{"E$<>$ ( chatbot.Ontology and chatbot2.TrainedBot)"}.
\textit{ii}) there is a system execution that puts the first instance "chatbot" in the TrainedBot state and the second instance "chatbot2" in the Ontology state using the formula \textbf{ "E$<>$ ( chatbot.TrainedBot and chatbot2.Ontology)"}. 
\textit{iii}) there is an execution of the system which allows to put the two instances in the Ontology state because, the chatbot has a read-only access to the ontology so two simultaneous accesses does not generate a problem \textbf{"E$<>$ ( chatbot.Ontology and chatbot2.Ontology)"}. We can clearly see from the image \ref{parallel ver} that all these rules are marked in green so validated. 

\begin {figure}[htbp]
  \hbox{ 
     \includegraphics[width=6cm, height=6cm]{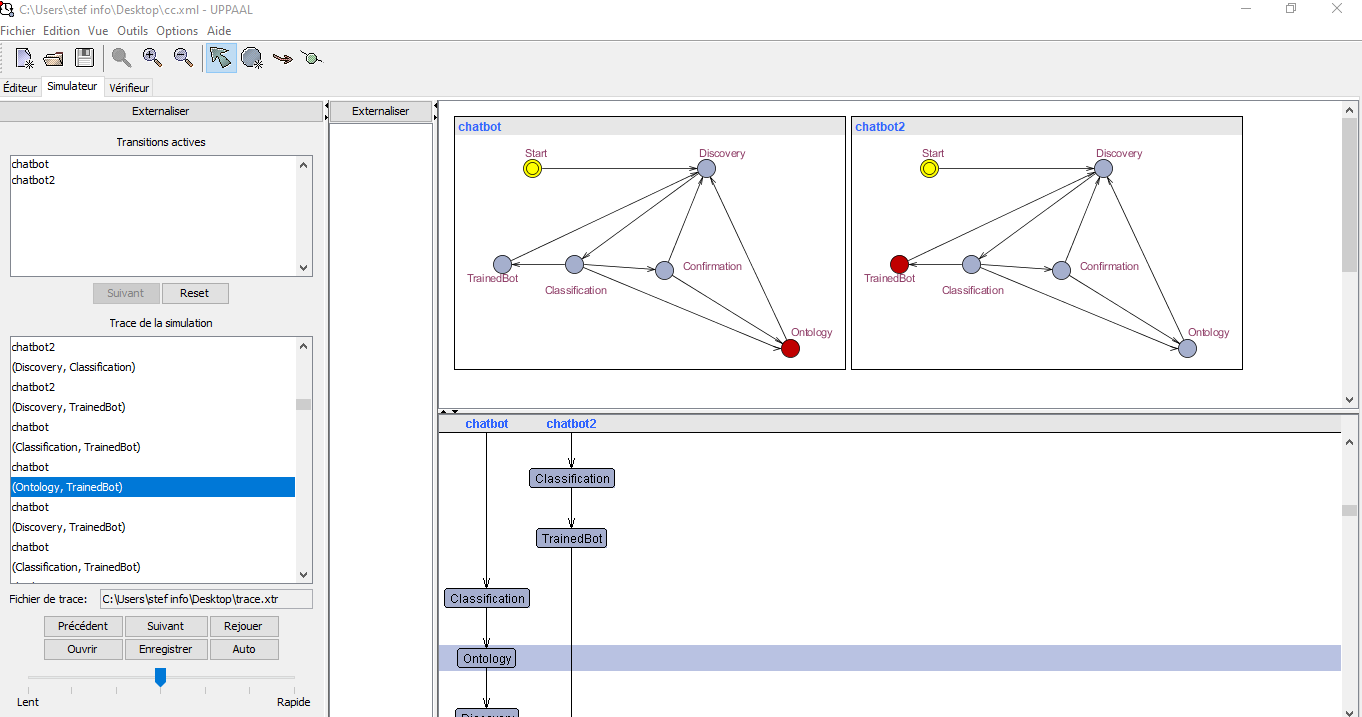}
     \hspace*{1cm}  
     \includegraphics[width=6cm,height=6cm]{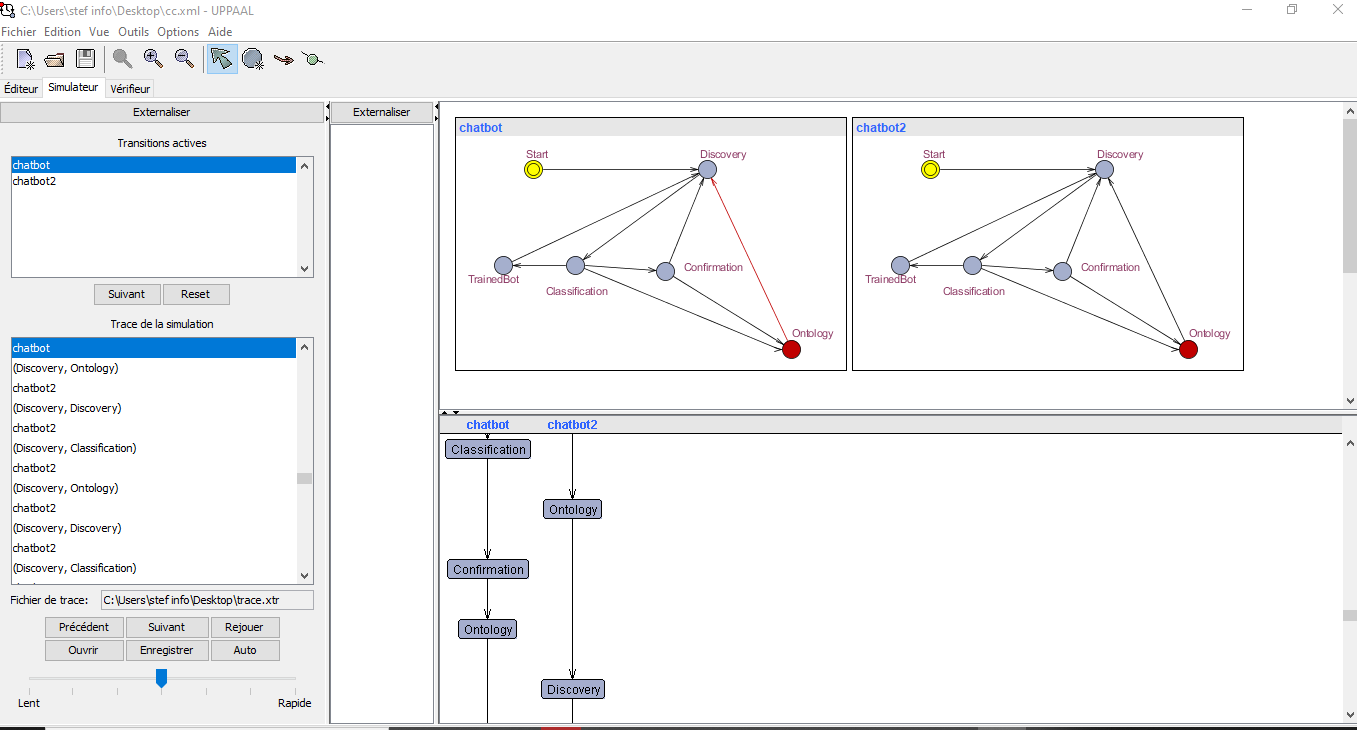}
  }
  \caption{Simulated execution}
  \label{simulateur}
\end {figure}

We will then access the simulator integrated into Uppaal, to see how the two instances of chatbot work. As shown in Figure \ref{simulateur}(1) the first instance is in the Ontology state and the second is in the TrainedBot state (this is the rule (\textit{i}) defined before). In Figure \ref{simulateur}(2) both instances of the system are in the Ontology state (this is the rule (\textit{iii})).

\newpage
\section{Conclusion}

In this chapter, We have illustrated the chatbot use by simulating some examples of conversations. Finally, we validate both the ontology and the state machine. The next chapter will summarize the content of all previous sections to recap what has been accomplished. It will give a global view of the completed system as well as providing pointers for future developments.

\chapter{Conclusion and futur work}
In this master thesis, we have made a chatbot with a new and innovative architecture which integrate several component, which are not integrated in this way before. The objective of our chatbot was to support citizens in emergency situations, and provide to them correct information and 24 hours of availability.

With the rapid advancement of technology, chatbots have become increasingly important in various domains. They are a big step forward in enhancing human computer interactions. Below I summarize the most important steps in the creation of our conversational agent.

\begin{enumerate}
    \item We created an ontology-driven chatbot; 
    \item We populated the ontology using web scraping technique; 
    \item Then we designed our use cases; 
    \item We implemented the state machine;
    \item We integrated a trained chatbot; 
    \item We created the data visualization module; 
    \item And finally we added the user interface. 
    
\end{enumerate}

Although all research has been done in a careful manner, some limitations must be taken into consideration, our future perspective to ameliorate our chatbot are described below.

\section{Futur work}
 
\subsubsection*{Universal chatbot:}
We aim to develop a universal chatbot, to reach this, in the future we will make it support many other languages.  
\subsubsection*{Make it smarter:}
To make our Covid-19 smarter we think about using more efficient NLP procedure, integrate a more sophisticated trained bot. Also we aim to apply voice recognition  techniques.

\subsubsection{More Informative}
To do so, the first step is to enrich the ontology with more information, such as trends and advises.

\nocite{ref1,ref2,ref3,ref4,ref5,ref6,ref7,ref8,ref9,ref10,ref11,ref12,ref13,ref14,ref15,ref16,ref17,ref18,ref19,ref20}

\bibliographystyle{plain}
\bibliography{main}

\newpage
\thispagestyle{empty}

\paragraph{Résumé \\ \\}

Aujourd'hui, c'est l'ère de l'intelligence dans les machines. Avec le progrès de l'intelligence artificielle, les machines ont commencé à imiter les humains, le chatbot est la prochaine grande chose dans le domaine des services de conversation. Un chatbot est une entité virtuelle capable de mener une conversation naturelle avec des personnes. Il peut inclure de compétences qui lui permettent de converser avec les humains sous forme audio, visuelle ou textuelle. Obtenir les bonnes informations au bon moment et au bon endroit est la clé d'une gestion efficace des catastrophes. Le terme "gestion des catastrophes" englobe à la fois les catastrophes naturelles et celles causées par l'homme. Pour aider les citoyens, notre projet est de créer Covid Assistant pour répondre au besoin d'informations actualisées qui doivent être disponibles 24 heures sur 24. Un chatbot peut être considéré comme un système de questions-réponses dans lequel des experts fournissent des connaissances pour solliciter les utilisateurs. Cette thèse de mastère est consacrée à la discussion du chatbot Covid Assistant et à l'explication détaillée de chaque élément. La conception du chatbot proposé est introduite par ses sept composantes : Ontologie, module de grattage Web, base de données, machine d'état, extracteur de mots-clés, un chatbot formé et interface utilisateur. \\

\textbf{Mots clés}:Chatbot, Agent conversationnel, Ontologie, Machine à Etat Fini, Gestion des catastrophes, Covid-19 \\ \\

\vspace{1pt}\hrule\vspace{4pt}

\end{document}